%% file: naacl-pretraining.tex
\pdfoutput=1
\documentclass[11pt]{article}
\usepackage[]{style/acl}

\usepackage{times}
\usepackage{latexsym}
\usepackage{multirow}
\usepackage[T1]{fontenc}
\usepackage{amsmath}
\usepackage{verbatim}
\usepackage{paralist}
\usepackage[utf8]{inputenc}

\usepackage{microtype}

\usepackage{inconsolata}
\usepackage{soul}

\usepackage{booktabs}
\usepackage{graphicx}
\usepackage[export]{adjustbox}
\usepackage{soul}
\usepackage{xcolor}

\newcommand{\hlg}[1]{#1}
\renewcommand{\hl}[1]{#1}

\usepackage{soulpos} % Extended version of `soul` for nested commands
\usepackage{etoolbox}

\newcommand{\tbd}[1]{}

\usepackage{colortbl}
\usepackage{pgfplots}

\usepackage{tabularx}
\usepackage{icomma}
\title{
  Making Language Models Robust Against Negation
}

\author{MohammadHossein Rezaei \and Eduardo Blanco \\
         Department of Computer Science, University of Arizona \\ \texttt{\{mhrezaei,eduardoblanco\}@arizona.edu}}

\begin{document}
\maketitle
\begin{abstract}
Negation has been a long-standing challenge for language models.
Previous studies have shown that they struggle with negation in many natural language understanding tasks.
In this work, we propose a self-supervised method to make language models more robust against negation.
We introduce a novel task, Next Sentence Polarity Prediction (NSPP), and a variation of the Next Sentence Prediction (NSP) task.
We show that BERT and RoBERTa further pre-trained on our tasks outperform the off-the-shelf versions on nine negation-related benchmarks.
Most notably, our pre-training tasks yield between 1.8\% and 9.1\% improvement on CondaQA, a large question-answering corpus requiring reasoning over negation.
\end{abstract}

\input{tex/introduction.tex}
\input{tex/related_work.tex}
\input{tex/empowering.tex}
\input{tex/reverse_polarity.tex}
\input{tex/dataset.tex}
\input{tex/evaluation.tex}
\input{tex/experimental_setup.tex}
\input{tex/results.tex}
\input{tex/conclusion.tex}

% \pagebreak
\section*{Limitations}
We experiment with two models, {RoBERTa} and {BERT}, and a single pre-training dataset, Wikipedia.
Future work may consider other models and pre-training datasets.
Our rules for reversing polarity only cover \hl{``not'', ``n't'', and ``never''}.
However, they are still effective in making models more robust against negation in general---recall that CondaQA has over 200 unique negation cues.
Future work may consider working with more sophisticated rules to reverse the polarity of the sentences.
We also only experiment with models pre-trained on 500K and 1M instances.
Future work may consider training on the whole corpus and evaluate the performance on downstream tasks.
Additionally, all the corpora we work with are in English.
We acknowledge that negation may be expressed differently in other languages and this work may not generalize to other languages.
We note, however, that the proposed tasks are language-agnostic and can be applied to other languages.

\section*{Ethics Statement}
The work in this paper does not involve human subjects.
We only use publicly available datasets and models.
We do not collect any personal information.
Therefore, this work does not raise any ethical concerns.

% Entries for the entire Anthology, followed by custom entries
\bibliography{custom}
% \bibliographystyle{acl_natbib}

%\clearpage
\appendix
\input{apptex/rules.tex}
\input{apptex/llm.tex}
\input{apptex/condaqa_example.tex}
\input{apptex/nlunli_example.tex}
\input{apptex/lama.tex}
\input{apptex/hyper.tex}
\input{apptex/ablation.tex}
\input{apptex/nli-details.tex}

\end{document}

%% file: tex/introduction.tex
\section{Introduction}
\label{sec:introduction}

Negation is a semantic phenomenon that alters an expression to convey the opposite meaning. 
It is a ubiquitous linguistic construct used in various forms across all languages.
Negation is present in approximately 25\% of English sentences \cite{hossain-etal-2020-analysis}.
Language Models (LMs) achieve remarkable performance in a wide range of natural language understanding tasks 
but have been shown to struggle when negation is involved 
\cite{ettinger-2020-bert,dobreva-keller-2021-investigating,hosseini-etal-2021-understanding,jang-etal-2022-beyond}.
\citet{truong-etal-2023-language} showed that Larger LMs 
such as GPT-3 
\cite{NEURIPS2020_1457c0d6}
and InstructGPT 
\cite{ouyang2022training} 
are also insensitive to the presence of negation and struggle to reason over it.
Although larger state-of-the-art models might be better at handling negation, 
it is still important to address this issue in smaller language models that are more computationally efficient and cheaper to deploy.

\begin{figure}
    \centering
    \input{tables/introexample.tex}
    \caption{
        An example of the training data for our self-supervised tasks.
        The tasks are: (a) given a sentence, predict whether the next sentence will contain negation~(NSPP)
        and (b) given two sentences, predict whether the second sentence is a coherent continuation of the first one~(NSP).
        \label{fig:introexample}
    }
\end{figure}

\citet{kassner-schutze-2020-negated} showed that {BERT}~\cite{devlin-etal-2019-bert}
often predicts the same token when negation is added to a sentence.
\hl{
For example, in the sentence \textit{A beagle is a type of [MASK]}, 
{BERT} predicts \textit{dog} as the masked token both in
the original sentence
and 
when the negation cue \textit{not} is added to the sentence,
\textit{A beagle is {not} a type of [MASK]}.}
We hypothesize that this behavior is due to the lack of negation modeling in pre-training.
Specifically, the model has not been exposed to instances where the addition (or removal) of negation \hlg{influences} meaning and coherence \hlg{within a discourse}.
We propose to further pre-train LMs on two novel tasks that involve negation. 
The first task is the Next Sentence Polarity Prediction (NSPP) task, 
where \hlg{given a sentence,} the model predicts whether the next sentence will contain negation. 
The second task is a variation of the well-known Next Sentence Prediction (NSP) task as introduced in {BERT}~\cite{devlin-etal-2019-bert}.
However, in our version, 
we \emph{reverse the polarity} of the second sentence \hlg{to form} negative samples rather than select random sentences from the document.
We define \emph{reversing polarity} as adding (or removing) negation to a sentence that contains (or does not contain) negation.
Figure~\ref{fig:introexample} shows \hlg{examples} of the self-supervised tasks.

The main contributions of this paper are:\footnote{All code and data available at \url{https://github.com/mhrezaei1/pretraining-negation} under the Apache 2.0 license.}
\begin{compactitem}
    \item Introducing \hl{two novel self-supervised tasks for pre-training LMs for negation.}
    \item Creating a large-scale dataset ($\approx$6.4M samples) for these tasks\hl{.}
    \item Showing that further pre-training {{BERT}}~\cite{devlin-etal-2019-bert} and {RoBERTa}~\cite{liu2019roberta} \hl{independently} on these tasks improves performance on CondaQA~\cite{ravichander-etal-2022-condaqa} and \hl{eight} other negation-related benchmarks. \hl{Joint pre-training on both tasks, however, does not always improve performance.}
\end{compactitem}

CondaQA, to the best of our knowledge, is the largest corpus that requires reasoning over negation. \hl{It consists of} $14,182$ question-answer pairs and over 200 unique negation cues.
Additionally, we also evaluate our models on three of the NLU corpora analyzed by~\citet{hossain-etal-2022-analysis} (QNLI~\cite{rajpurkar-etal-2016-squad},
WiC~\cite{pilehvar-camacho-collados-2019-wic},
and 
WSC~\cite{levesque_winograd_2012}),
and the new negation benchmarks introduced by
\citet{hossain-etal-2020-analysis}
for RTE~\cite{dagan2005pascal},
SNLI~\cite{bowman-etal-2015-large},
and MNLI~\cite{williams-etal-2018-broad}.
Finally, we evaluate on LAMA~\cite{petroni-etal-2019-language}
and LAMA-neg~\cite{kassner-schutze-2020-negated}.\tbd{No paper is being cited more than once here.}

%% file: tables/introexample.tex
\small
\centering
\begin{tabularx}{0.49\textwidth}{l@{\hspace{0.1cm}}X}
\toprule
Sentence ($S_1$): & {The computer screen stayed blank.}  \\
\midrule
Next Sentence & \\
~Original ($S_2$): & {It {didn't} display any images.} \\
~Reversed Polarity ($S_2'$): & {It displayed some images.} \\
\midrule
\midrule
\multicolumn{2}{l}{Next Sentence Polarity Prediction (NSPP)} \\
\multicolumn{2}{l}{~~~~Input: $S_1$} \\
\multicolumn{2}{l}{~~~~Label: {Yes}, the sentence after $S_1$ (i.e., $S_2$) has negation.} \\
\midrule
\multicolumn{2}{l}{Next Sentence Prediction (NSP)} \\
\multicolumn{2}{l}{~~~~Input: ($S_1$, $S_2$), Label: {Yes}, $S_2$ is the next sentence.} \\
\multicolumn{2}{l}{~~~~Input: ($S_1$, $S_2'$), Label: {No}, $S_2'$ is not the next sentence.} \\
\bottomrule
\end{tabularx}

%% file: tex/related_work.tex
\section{Related Work}
\label{sec:related_work}

Negation understanding studies have focused on various aspects of negation, 
including negation detection~\cite{khandelwal-sawant-2020-negbert,truong-etal-2022-improving,morante-daelemans-2012-conandoyle},
scope detection~\cite{qian-etal-2016-speculation,jumelet-hupkes-2018-language,fancellu-etal-2017-detecting,morante-daelemans-2009-metalearning,li-lu-2018-learning,zhao-bethard-2020-berts}, and
focus detection~\cite{rosenberg-bergler-2012-uconcordia,shen-etal-2019-negative,zou-etal-2014-negation,zou-etal-2015-unsupervised}.

Recent studies have also focused on improving the performance of LMs when negation is present.
\citet{sarabi-etal-2019-corpus} introduced \emph{affirmative interpretations}, 
i.e., paraphrasing negated sentences without using negation cues.
\citet{hossain-blanco-2022-leveraging}
and 
\citet{rezaei2024paraphrasing}
showed that incorporating affirmative interpretations improves the performance of LMs
on negation-related benchmarks. 
In this paper, we show that our \hl{pre-training strategies} outperform affirmative interpretations on CondaQA. 
Unlike theirs, our \hl{pre-training} does not add any complexity at prediction time.

\citet{hosseini-etal-2021-understanding} propose to augment negation understanding with {BERT} by unlikelihood training and synthetic data generation.
\citet{singh-etal-2023-nlms} repeats the pre-training procedure of {BERT} (i.e., MLM and NSP) modified to improve negation understanding.
They add negation cues to the original next sentences rather than choosing a random sentence.
Unlike them, we do not always add negation; instead, we reverse the polarity of the second sentence for the NSP task.
That is, half the time we add negation to the next sentence and half the time we remove it.
We also introduce a novel task, 
Next Sentence Polarity Prediction (NSPP),
which is not explored by previous work.
As we shall see,
our pre-training outperforms previous work.

%% file: tex/empowering.tex
\section{Empowering Language Models Against Negation}
\label{sec:empowering}

We propose a self-supervised method to make LMs more robust against negation.
Our approach is to further pre-train LMs on two tasks that involve negation.
These tasks are the Next Sentence Polarity Prediction (NSPP) task 
and a variation of the well-known Next Sentence Prediction (NSP) task.
None of these tasks require labeled data\hl{;
any text corpora are suitable.}
Also, they are not specific to any domain or downstream task.

\subsection{Next Sentence Polarity Prediction (NSPP)}
\label{sec:nspp}
We introduce NSPP as the task of predicting the polarity of the next sentence given the current sentence.
\hlg{Given} a pair of consecutive sentences, $(S_1, S_2)$, 
the input to the model is only $S_1$, 
and the output is a binary label indicating whether $S_2$ \hlg{includes} any negation cues or not.
For example, consider the following pair of sentences:
\begin{compactitem}
  \item[] $S_1$: \hl{\textit{The weather report showed sunny skies.}}% \textit{``I like apples.''}
  \item[] $S_2$: \hl{\textit{But it didn’t stay that way.}}% \textit{``I don't like oranges.''}
\end{compactitem}
Given only $S_1$,
the model should predict that the following sentence \hlg{includes} negation cues.

\subsection{Next Sentence Prediction (NSP)}
\label{sec:nsp}
NSP is a well-known task in LM pre-training as first introduced by BERT~\cite{devlin-etal-2019-bert}.
The NSP task is to predict whether two sentences are consecutive.
\citet{devlin-etal-2019-bert} (a) used consecutive sentences from Wikipedia as positive examples 
and (b) chose a random sentence from the same article to replace the second sentence \hl{and} create a negative example.

We propose a variation of the NSP task to improve negation understanding.
For a pair of consecutive sentences, $(S_1, S_2)$,
we create the negative pair $(S_1, S_2')$ 
where $S_2'$ is obtained by reversing the polarity of $S_2$. 
That is, if $S_2$ includes negation cues, we remove them, 
and vice versa. 

%% file: tex/reverse_polarity.tex
\subsubsection{Reversing Polarity}
\label{sec:reversing}

We define rules to add and remove negation cues from sentences.
These rules are used to create the negative pairs $(S_1, S_2')$ in the NSP task.
To streamline the process,
we only work with sentences that
\begin{compactitem}
    \item include \emph{not}, \emph{n't}, or \emph{never} as negation cues;
    \item the negation cue modifies the main verb;
    \item are not questions; and
    \item contain exactly one negation cue.
\end{compactitem}
To develop the rules, 
we collected a large set of sentences from the English Wikipedia corpus~\cite{wikidump} that met these criteria.
We then generated the dependency tree for each sentence with spaCy~\cite{spacy2, honnibal-johnson-2015-improved} 
and analyzed the frequency of outgoing edges from the main verb. 
Afterward, we manually inspected the most frequent tokens associated with each edge 
and leveraged these patterns to develop the \hl{rules below}.

\hl{
    We evaluated these rules by manually inspecting 100 samples.
    In 96\% of them, 
    the rules correctly reverse polarity. 
    Note that the goal here is not 100\% correctness---it is to automatically generate data for pre-training with our tasks.
} \tbd{Is em dash (---) correct here?}

\paragraph{Adding negation.}
For sentences where the main verb has no auxiliary verb,
we insert the negation cue directly and adjust the verb for tense and subject agreement. 
The cue \emph{never} is always placed directly before the main verb.
We append \emph{n't} and \emph{not} directly after the main verb if it is one of the following:
\emph{were}, \emph{was}, \emph{is}, \emph{are}, \emph{do}, \emph{will}, \emph{would}, \emph{may}, \emph{might}, \emph{shall}, \emph{should}, \emph{can}, \emph{could}, or \emph{must}.
For example, given the sentence \emph{``I was shopping.''},
we add \emph{not} to create the sentence \emph{``I was not shopping.''}.

If the main verb is a gerund or present participle, we do not add \emph{n't} directly to it; 
instead, we place \emph{not} right before the verb. 
For present or past participles,
we replace it with its lemma and insert the appropriate form of \emph{do} before the lemma, 
ensuring it matches the tense of the verb and person of the subject.
For present participles, we add \emph{do} or \emph{does}, and for past participles, we add \emph{did}.
We then insert \emph{not} or \emph{n't} after the auxiliary verb.
For example, given the sentence \emph{“I went to the store,”} 
the main verb \emph{went} is replaced with \emph{did not go}, resulting in \emph{“I did not go to the store.”}

If the main verb has an outgoing edge labeled \emph{aux} or \emph{auxpass} in the dependency tree,
we add the negation cue to the auxiliary verb.
For example, given the sentence \emph{``The store is closed,''}
we add \emph{n't} to the auxiliary verb \emph{is} to create the sentence \emph{``The store isn't closed.''}
\hl{
    However, for certain auxiliary verbs such as \emph{might} and \emph{may},
    it is not possible to add \emph{n't} directly to them.
    In such cases, we only add \emph{not} or \emph{never} to the sentences.}
    Appendix~\ref{app:reversing} \hl{lists the auxiliary verbs we work with and the rules for adding each negation cue.

    Additionally, to have more natural sentences with negation cues,
    we replace modifiers such as \emph{already} and \emph{some} with \emph{yet} and \emph{any}, respectively.
}

\paragraph{Removing negation.}
We begin by removing the negation cue from the sentence and adjusting the grammar accordingly.
If the negation cue is \emph{n't} (as in \emph{can't} or \emph{won't}), 
we remove \emph{n't} and replace the auxiliary verb with its lemma (e.g., \emph{can} and \emph{will}).

{Next, we remove any extra auxiliary verbs and adjust the main verb based on tense and subject agreement. 
If the auxiliary verb is \emph{did}, we remove \emph{did} and use the past tense form of the main verb.
For example, given the sentence \emph{``I did not go to the store,''} we remove \emph{did} and update \emph{go} to \emph{went}, 
resulting in \emph{``I went to the store''}. 
We apply the same process for \emph{do} and \emph{does}. 
That is, we replace the main verb with its base form or third-person singular form, respectively.
}

We also replace negative polarity items such as \emph{yet}, \emph{at all}, and \emph{any}
with their affirmative counterparts (\emph{already}, \emph{somewhat}, and \emph{some}, respectively.)
Lastly, 
if \emph{but} functions as a conjunction and is a sibling of the main verb in the dependency tree, we replace it with \emph{and}.

\paragraph{A note on using LLMs.}
Although using LLMs is expensive and time-consuming, 
we attempted to use state-of-the-art LLMs to reverse the polarity of sentences.
We used the Llama-2 model \cite{touvron2023llama} and the GPT-4 model \cite{openai2024gpt4}. 
We tried several prompting approaches 
to instruct the models to only \hl{add or remove} negation cues \hl{without modifying} other parts of the sentence.
However, the models \hl{consistently made additional modifications to keep the meaning of the sentence intact. } 
\hl{
    We hypothesize that this is because we work with Wikipedia sentences, which are typically about facts.
    Since these models are believed to be trained to be truthful, 
    they often refuse to generate text that contradicts real-world facts.
}
See examples of the prompts and outputs in Appendix~\ref{app:llm-reversing}.

%% file: tex/dataset.tex
\subsection{Dataset}
\label{sec:dataset}

We begin by extracting all sentences \hl{from Wikipedia} containing negation that are not the first sentence of a section, 
ensuring that each has a preceding sentence ($S_1$) to provide context for \hl{the next sentence} ($S_2$).
Next, we retain only the sentences where our rules for reversing polarity can be applied.
We also extract affirmative sentences (i.e., without negation cues) along with their preceding sentences ($S_1$).
To ensure the sentences are truly affirmative, we verify that none of the negation cues from CondaQA are present.
Additionally, 
we aim to select the same number of affirmative sentences from the same article as the negated sentences.
If this is not possible, the remaining sentences are drawn from a different article.
We obtain an equal number of negated and affirmative sentences, resulting in a total of $\approx$6.4M $(S_1, S_2)$ pairs. 
The negated and affirmative sentences are used as the positive and negative examples for the NSPP task, respectively.
Next, we create $(S_1, S_2')$ pairs by reversing the polarity of the $S_2$ sentences.
We add only one negation cue to the affirmative sentences
and ensure that we have the same distribution than in the negated sentences for each article.

We randomly choose $25,000$ ($S_1, S_2$) pairs along with their corresponding $(S_1, S_2')$ pairs to create the validation sets for pre-training.
Half of the $S_2$ sentences are negated and the other half are affirmative.
The NSPP task does not need the ($S_1, S_2'$) pairs.
Therefore, the validation sets consists of $25,000$ and $50,000$ pairs for the NSPP and NSP tasks, respectively.

%% file: tex/evaluation.tex
\section{Evaluation Benchmarks}
\label{sec:evaluation}

We evaluate off-the-shelf LMs and versions further pre-trained with our strategies on benchmarks that require reasoning over negation.
We primarily evaluate on CondaQA~\cite{ravichander-etal-2022-condaqa} \hl{and provide results on eight other relevant benchmarks.}
All the corpora that we work with are in English.

\subsection{CondaQA}
\label{sec:condaqa}

CondaQA \cite{ravichander-etal-2022-condaqa} is a contrastive question-answering dataset.
It is designed to evaluate the ability of models to reason over negation.
The corpus consists of $14,182$ question-answer pairs. 
Each question is paired with a passage that contains the answer to the question.
Answers to questions are either Yes/No/Don't Know, a span in the question, or a span in the passage.
It was created by extracting $1,289$ passages from the English Wikipedia 
which contained negation phrases.
They started with a collection of negation cues 
\cite{Morante2011NegationCues, van-son-etal-2016-building} 
and expanded them. 
There are over 200 unique negation cues in CondaQA,
including 
single-word (e.g., not, never), 
\hl{affixal} (e.g., \emph{un}-lucky, \emph{in}-correct),
and multi-word negation cues (e.g., a lack of, instead of).
\hl{They also include} several parts of speech tags such as
verbs (e.g., refuse), 
nouns (e.g., absence), 
adjectives (e.g., uninterested), 
and adverbs (e.g., incorrectly).

\hl{The authors of CondaQA} instructed crowdworkers to make three types of edits to each passage:
\begin{compactitem}
    \item {Paraphrase}: Rewrite the negated sentence.
    \item {Scope}: Change the scope of the negation, i.e., the part of the sentence that is negated.
    \item {Affirmation}: Remove the negation from the sentence.
\end{compactitem}

The \emph{paraphrase} edit keeps the meaning unchanged but \hl{the} \emph{scope} and \emph{affirmation} edits change the meaning of the sentence.
The questions are written by crowdworkers to target the implication of the negated statement in the passage.
The questions are then answered for the original and edited passages. 
Appendix~\ref{sec:condaqaexample} shows examples.

Importantly, \citet{ravichander-etal-2022-condaqa} {demonstrated that CondaQA cannot be solved by models relying solely on questions, edit types, or cues. 
This ensures that the dataset is free from artifacts that models could exploit to solve the task.}

We evaluate our models on the CondaQA dataset using the accuracy and \emph{group consistency} metrics.
The term \emph{group} refers to the original passage and either all three or one of the edited passages.
\emph{Group consistency} measures the percentage of questions answered correctly for all the passages in a group.
Group consistency is arguably more important,
as being robust against negation means being able to answer the question correctly with all original and edited passages.

\begin{figure*}
    \centering
    \includegraphics[width=0.32\textwidth]{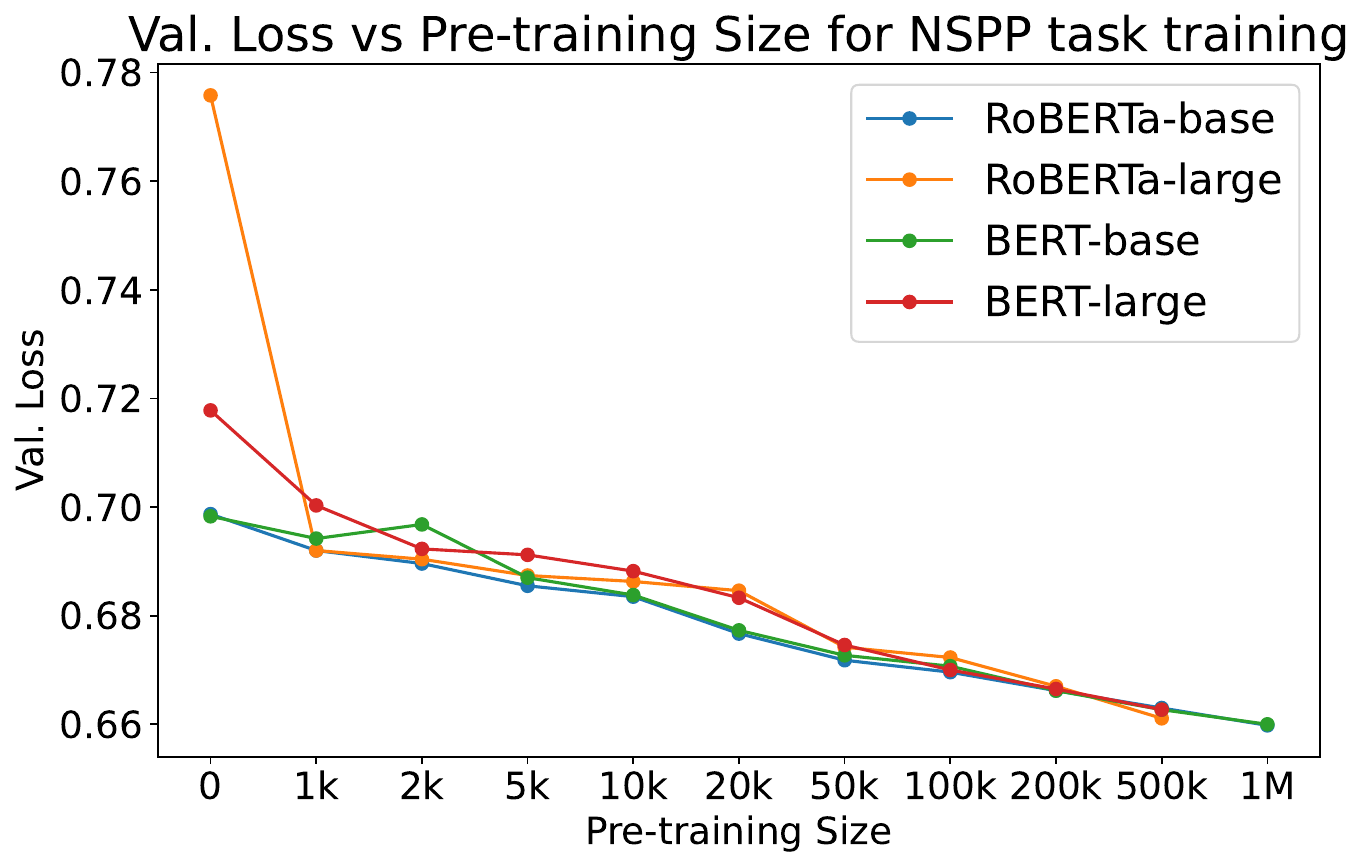}
    \includegraphics[width=0.32\textwidth]{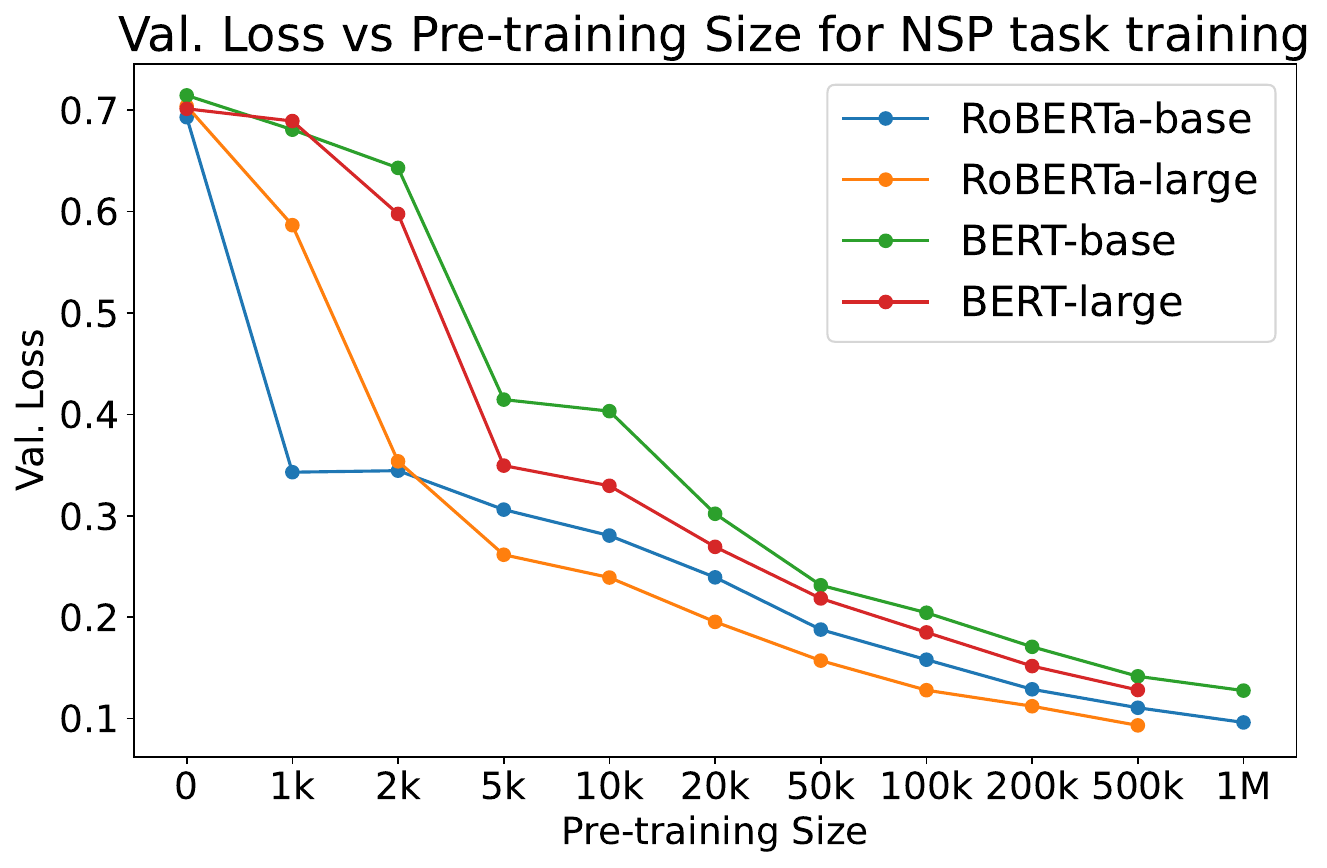}
    \includegraphics[width=0.32\textwidth]{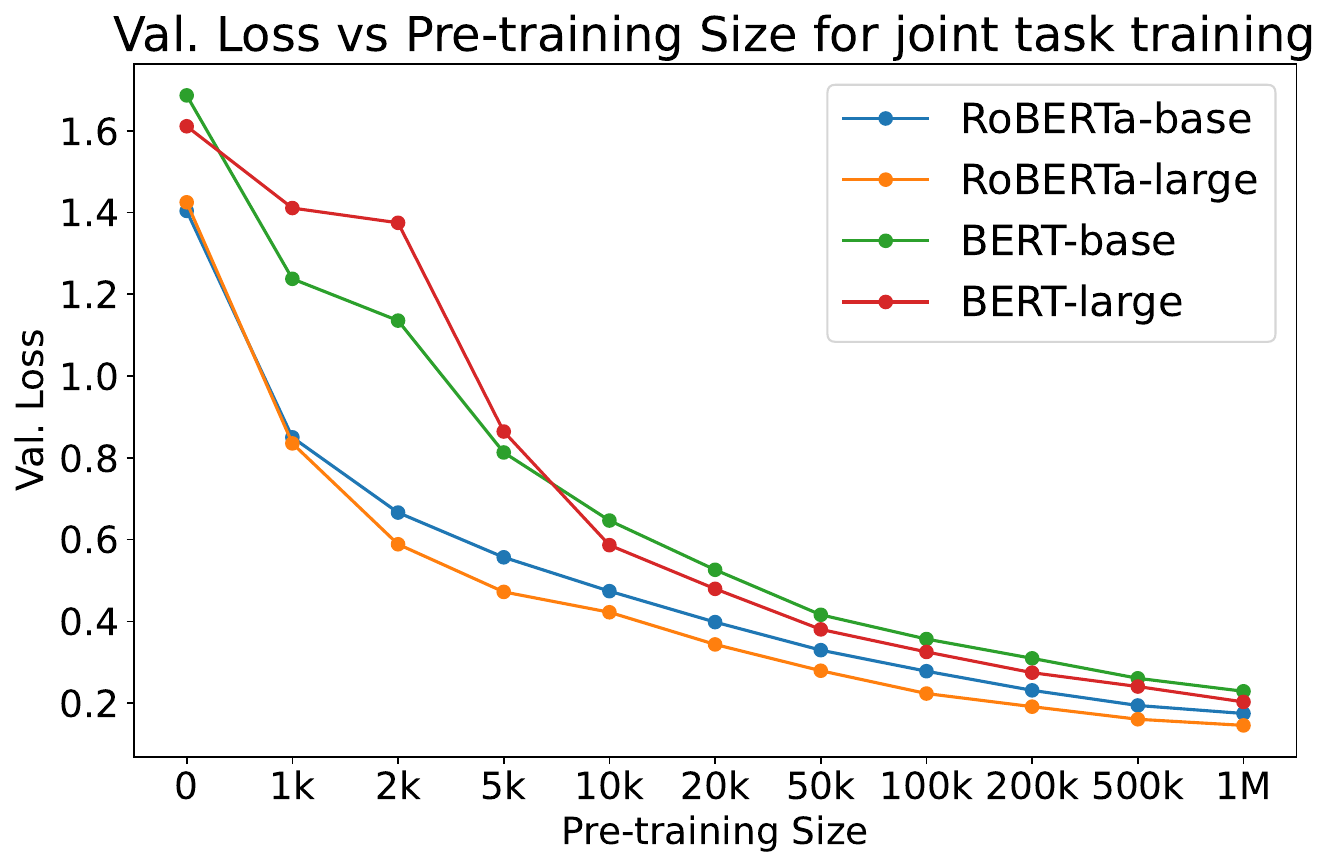}
    \caption{
        Trends in \hl{pre-training} transformers on NSPP, NSP, and both tasks jointly from left to right.
        Validation loss decreases as the model is trained on larger subsets of the corpus.
        We stop training when the validation loss plateaus.
        \label{fig:pretraining}
    }
\end{figure*}
\subsection{Other Corpora}
\label{sec:othercorpora}
\paragraph{NLI and NLU Corpora.}
\citet{hossain-etal-2020-analysis} and \citet{hossain-etal-2022-analysis} \hl{analyzed negation in existing NLI and NLU corpora.
They showed that there are few negation cues in these corpora,
and the cues are often \emph{unimportant} for the task.
That is, negation can be removed from the sentences without changing the label of the example.}

Furthermore,
they introduced three new negation benchmarks built from existing NLI corpora.
They worked with RTE~\cite{dagan2005pascal,BarHaim2006TheSP,giampiccolo-etal-2007-third,DBLP:conf/tac/BentivogliMDDG09},
SNLI~\cite{bowman-etal-2015-large}, and MNLI~\cite{williams-etal-2018-broad}.
\hl{To create their benchmarks, they randomly selected 500 text and hypothesis pairs from each corpus.
They then added negation cues to the main verb of the text and hypothesis sentences to create three new pairs from each original pair.
The new pairs were manually annotated using the labels of the original benchmarks.}
Appendix~\ref{sec:nlunlicorpora} contains more details and examples from the NLU and NLI corpora that we work with.

\paragraph{LAMA and LAMA-Neg.}
LAMA probe~\cite{petroni-etal-2019-language} is a benchmark for evaluating the factual and commonsense knowledge of language models.
LAMA is composed of various datasets, namely
GoogleRE,\footnote{\footnotesize {code.google.com/archive/p/relation-extraction-corpus/}}~T-REx~\cite{elsahar-etal-2018-rex}, ConceptNet~\cite{speer-havasi-2012-representing}, and SQuAD~\cite{rajpurkar-etal-2016-squad}. 
The input to the model is a sentence with a masked token.
The goal is to correctly predict the masked token.
LAMA-Neg~\cite{kassner-schutze-2020-negated} is a variant of LAMA where negation cues are inserted into the sentences.
The model is supposed to predict any token other than the original token. 
Appendix~\ref{app:lama} provides more information about these corpora.

%% file: tex/experimental_setup.tex
\section{Experiments}

First, we further pre-train transformers on the tasks described in Section~\ref{sec:empowering}.
Then, evaluate the models on the benchmarks described in Section~\ref{sec:evaluation}.
    
\subsection{Pre-training LMs for Negation}
We use base and large versions of {BERT}~\cite{devlin-etal-2019-bert} and {RoBERTa}~\cite{liu2019roberta} as our baseline models.
We further pre-train the models on NSPP and NSP tasks individually and jointly.
Since \hl{{BERT} and {RoBERTa}} are already pre-trained on Wikipedia using masked language modeling, further masked language modeling during our pre-training is redundant.
We use Transformers ~\cite{wolf-etal-2020-transformers}
and PyTorch~\cite{paszke2019pytorch} libraries.
We use the Adam optimizer~\cite{kingma2017adam} with a learning rate of $1e-6$
and Cross-Entropy loss for both tasks. 
We use early stopping with a patience of 3 epochs on the validation set.

For joint training on NSPP and NSP tasks, we use the same learning rate and optimizer. 
The output of the model's last layer is fed into two separate linear layers,
each predicting the label for its respective task. 
The total loss is computed as the sum of the individual task losses. 
The input to the model remains the same as in the NSP task, 
with two sentences concatenated using the special token [SEP].
When pre-training only on the NSPP task, the input is only the first sentence $S_1$.
However, 
the presence of the second sentence in the input when pre-training jointly does not make the task trivial.
Note that in the NSP task, we reverse the polarity of the second sentence half of the time.
Therefore, even if the model looks for negation cues in the second sentence,
it cannot rely on them to predict the polarity of the next sentence in the NSPP task.
\begin{table*}[t]
    \centering
    \input{tables/condaqaresults.tex}
    \caption{
        Results on CondaQA test set.
        The first group of rows shows the results from previous work.
        The second group of rows shows the results of \hl{further pre-training on our tasks}.
        An asterisk ($^{\ast}$) indicates a statistically significant improvement (McNemar's test \cite{mcnemar1947note}, $p < 0.05$)
        over \hl{off-the-shelf} version of LMs.
        Pre-training {BERT} and {RoBERTa} on any of our tasks statistically significantly outperforms \hl{off-the-shelf} LMs.
        {RoBERTa-large} pre-trained on any of the tasks achieves higher group consistency (and accuracy)
        than (1) using affirmative interpretations
        and (2) {UnifiedQA-v2-large}\hl{,} which has been pre-trained on $\approx$1M annotated question-answer pairs.
        \label{tab:condaqaresults}
    }
\end{table*}

\hl{Training} on the entire corpus
($\approx$12.8M sentence pairs for NSP and $\approx$6.4M for NSPP)
is computationally expensive and time-consuming. 
Instead, we train on smaller subsets of the corpus, 
gradually increasing their sizes as training progresses, 
until the validation loss plateaus.
Figure~\ref{fig:pretraining} \hl{plots losses as more pre-training data is used. }
We observe similar trends for {BERT} and {RoBERTa} models across both base and large configurations.
We stop pre-training after 1M sentence pairs, 
except for the large models trained on NSP and NSPP tasks, 
where we stop after 500K pairs. 
While the loss reduction on the NSPP task is not as substantial as on the NSP task, 
the model still demonstrates improved performance on downstream tasks.

\subsection{Downstream Tasks}
For the downstream tasks (other than LAMA that has no training set),
we further fine-tune the pre-trained models on the benchmarks described in Section~\ref{sec:evaluation}.
More details on the implementation and hyperparameters are provided in Appendix~\ref{app:hyperparameters}.

%% file: tables/condaqaresults.tex
\newcommand{\sig}{$^{\ast}$}
\setlength{\tabcolsep}{0.07in}
\small
\begin{tabular}{l r c@{\hspace{0.10in}}  r@{}l r@{}l@{\hspace{0.07in}} r@{}l@{\hspace{0.07in}} r@{}l@{\hspace{0.07in}} r@{}l@{\hspace{0.07in}}   c@{\hspace{0.10in}}   r@{}l r@{}l r@{}l r@{}l}
\toprule
& & & \multicolumn{10}{c}{Accuracy} && \multicolumn{8}{c}{Group Consistency} \\
\cmidrule{4-12} \cmidrule{15-22}
& \# Params. && All && Ori. && Par. && Sco. && Aff. &&& All && Par. && Sco. && Aff. \\
\midrule
From \citet{ravichander-etal-2022-condaqa} & \\
~~~~{RoBERTa-large} & 355M && 54.1 && n/a && n/a && n/a && n/a &&& 13.6 && 51.6 && 26.5 && 27.2 & \\
~~~~{UnifiedQA-v2-base} & 220M && 58.0 && n/a && n/a && n/a && n/a &&& 17.5 && 54.6 && 30.4 && 33.0 & \\
~~~~{UnifiedQA-v2-large} & 770M && 66.7 && n/a && n/a && n/a && n/a &&& 30.2 && 64.0 && 43.7 && 46.5 & \\
~~~~{UnifiedQA-v2-3B} & 3B && 73.3 && n/a && n/a && n/a && n/a &&& 42.2 && 72.8 && 55.7 && 57.2 & \\

From \citet{rezaei2024paraphrasing} & \\
\multicolumn{1}{l}{~~~~{RoBERTa-large} + Affir.}& 355M && 67.1 && 68.5 && 68.0 && 61.8 && 69.7 &&& 31.4 && 61.9 && 43.8 && 50.7 & \\
\midrule
\midrule
\multicolumn{4}{l}{\hl{Our study}} & \\

~~~~{BERT-base} & 110M && 45.4 && 44.8 && 44.7 && 45.0 && 47.3 &&& 3.4 && 40.5 && 15.2 && 14.1 \\
~~~~~~~~ + NSPP &&& 47.2 &\sig& 48.7 &\sig& 48.1 &\sig& 46.3 && 45.7 &&& 2.7 && 45.9 && 16.3 && 12.0 \\
~~~~~~~~ + NSP &&& \textbf{47.8} &\sig& 48.8 &\sig& 49.4 &\sig& 45.9 && \textbf{46.9} &&& \textbf{3.6} && 44.2 && 16.8 && \textbf{13.7} \\
~~~~~~~~ + NSPP + NSP &&&  47.8 &\sig& \textbf{49.8} &\sig& \textbf{50.0} &\sig& \textbf{47.6} && 43.7 &&& 2.9 && \textbf{46.3} && \textbf{17.2} && 12.1 \\
\midrule
~~~~{BERT-large} & 336M && 48.5 && 49.4 && 49.6 && 47.0 && 47.8 &&& 3.6 && 45.7 && 16.5 && 13.6 \\
~~~~~~~~ + NSPP &&&  51.8 &\sig& 51.8 && 51.0 && 49.7 && \textbf{54.5} &\sig&& 8.4 && 41.1 && 23.3 && 24.0 \\
~~~~~~~~ + NSP &&&  \textbf{51.9} &\sig& \textbf{53.3} &\sig& 51.0 && \textbf{50.5} &\sig& 52.7 &\sig&& \textbf{11.1} && 45.2 && \textbf{25.5} && \textbf{25.5} \\
~~~~~~~~ + NSPP + NSP &&&  50.3 &\sig& 52.0 &\sig& \textbf{51.9} &\sig& 48.4 && 48.6 &\sig&& 4.9 && \textbf{46.5} && 19.8 && 16.3 \\

\midrule
~~~~{RoBERTa-base} & 125M && 48.2 && 50.1 && 49.9 && 48.5 && 44.1 &&& 2.4 && 48.4 && 17.0 && 11.1 \\
~~~~~~~~ + NSPP &  &&  56.1 &\sig&  56.2 &\sig&  55.4 &\sig&  53.7 &\sig&  59.0 &\sig&&  16.9 && 49.5 &&  30.1 &&  33.1 \\
~~~~~~~~ + NSP &  &&  \textbf{57.3} &\sig&  57.8 &\sig& \textbf{57.9} &\sig&  \textbf{53.7} &\sig&  \textbf{59.6} &\sig&&  \textbf{17.6} &&  \textbf{50.4} &&  \textbf{30.8} &&  \textbf{34.6} \\
~~~~~~~~ + NSPP + NSP &  &&  56.7 &\sig&  \textbf{59.7} &\sig&  56.4 &\sig&  53.2 &\sig&  57.4 &\sig&&  15.6 &&  50.4 &&  29.8 &&  34.1 \\
\midrule
~~~~{RoBERTa-large} & 355M && 64.9 && 66.6 && 66.3 && 60.7 && 65.6 &&& 29.6 && 61.9 && 41.4 && 45.8 \\
~~~~~~~~ + NSPP &  &&  67.3 &\sig&  69.7 &\sig&  69.1 &\sig&  62.9 &&  67.0 &&&  33.5 &&  64.6 &&  46.3 &&  50.2 \\
~~~~~~~~ + NSP &  &&  68.0 &\sig&  \textbf{71.1} &\sig&  \textbf{70.2} &\sig&  63.3 &\sig&  67.1 &&&  34.0 &&  \textbf{67.0} &&  46.6 &&  50.1 \\
~~~~~~~~ + NSPP + NSP &  &&  \textbf{68.3} &\sig&  69.7 &\sig&  69.8 &\sig&  \textbf{64.8} &\sig&  \textbf{68.7} &\sig&&  \textbf{34.8} &&  64.4 &&  \textbf{47.0} &&  \textbf{50.5} \\
\bottomrule

\end{tabular}

%% file: tex/results.tex
\section{Results and Discussion}
\label{sec:results}

We first present the results on CondaQA and then on the \hl{eight} other corpora introduced in Section~\ref{sec:evaluation}.

\subsection{Results on CondaQA}
\label{sec:condaqaresults}

Table~\ref{tab:condaqaresults} presents the results with the CondaQA test set.
{BERT} and {RoBERTa} models further pre-trained on any of our tasks consistently achieve 1.8\%-9.3\% higher accuracy compared to the \hl{off-the-shelf} models. 
Importantly, \hl{all} improvements are statistically significant. 
\hl{Group consistency,
which according to} \citet{ravichander-etal-2022-condaqa} \hl{is a more reliable metric for evaluating the ability of models to reason over negation,
also improves for all models.}
The only exception is {BERT-base} pre-trained on NSPP or both tasks jointly where the difference is marginal (2.7\% and 2.9\% vs. 3.4\%).
Most notably, {RoBERTa-base} pre-trained on any of the tasks achieves between 15.6\% and 17.6\% group consistency compared to 2.4\% of off-the-shelf.

\begin{table*}[h!]
    \centering
    \input{tables/nliresults.tex}
    \caption{
        \hl{Results on the natural language inference tasks.
        We report accuracy on the development sets (dev) and the subset of the development sets that contain negation ($\text{dev}_{\text{neg}}$).
        Additionally, we evaluate the models on the new pairs created by }\citet{hossain-etal-2020-analysis}\hl{ that add negation to the main verb of the text and hypothesis sentences (Negated).}
        Further pre-training on NSP consistently outperforms the \hl{off-the-shelf} models on the new negated pairs (except for {RoBERTa-base} on SNLI and MNLI).
        It also outperforms the previous work other than {RoBERTa} on RTE.
        \label{tab:nliresults}
    }
\end{table*}
Recall that \hl{the} affirmation edit of the negated sentences is obtained by undoing negation (i.e., removing negation cues),
\hl{then the edited} sentence is affirmative.
\hl{We observe that the same improvements on accuracy and group consistency hold with the affirmation edits as well} (except for the {BERT-base} models).
This suggests that our proposed pre-training tasks enhance model robustness beyond just reasoning over negation.

{UnifiedQA-v2} \cite{khashabi2022unifiedqav2} models outperform \hl{off-the-shelf} {RoBERTa} models. 
However, they have been fine-tuned on $\approx$1M question-answering pairs. 
Our models, in contrast, have not been pre-trained on any question-answering data \hl{yet} outperform {UnifiedQA-v2} with\hl{ similar amount} of parameters.
All {RoBERTa-large} \hl{(355M)} models pre-trained on our tasks outperform the large (770M) {UnifiedQA-v2} model by 3.3\%-4.6\% in group consistency and 0.6\%-1.6\% in accuracy.

\begin{table*}[t]
    \centering
    \input{tables/nluresults.tex}
    \caption{
        Macro-averaged F1 scores on the validation sets of natural language understanding benchmarks.
        We report the performance on the original instances and the subset of instances that contain negation.
        All instances with negation in WiC and WSC are unimportant (i.e. can be answered without considering negation).
        The main takeaways are: (a) pre-training on NSP consistently improves the performance on instances with negation, and
        (b) pre-training on NSPP or joint pre-training is also either beneficial or does not substantially change the performance.
        \label{tab:nluresults}
        }
\end{table*}

Additionally,
we compare our models with the previous work by \citet{rezaei2024paraphrasing}.
They automatically generate affirmative interpretations of the negated sentences and add them to the input.
However, pre-training on any of our tasks is more effective than coupling the input with affirmative interpretations
(2.1\%--3.4\% in group consistency and 0.2\%--1.2\% improvement in accuracy). 
\hl{Importantly, }
our method bypasses the need to generate affirmative interpretations of the negated sentences.

\hl{
While pre-training on any of the tasks is beneficial,
we observe that pre-training on NSP consistently outperforms pre-training on NSPP.
Jointly training on both tasks improves the performance for the RoBERTa-large model by 0.3\% in accuracy and 0.8\% in group consistency.
However, it decreases the performance for the rest of the models. 
This is likely because the NSP task is easier to learn as it requires the model to predict the coherence of the two sentences in the input.
The NSPP task, on the other hand, is more challenging and requires the model to predict the polarity of a sentence solely based on the previous sentence.
We hypothesize that when trained jointly, the NSPP task introduces complexity that interferes with the model's ability to effectively learn tasks and reason over negation.
}

Additionally, 
we ran an ablation experiment where we only added or removed negation cues in the pre-training data.
Appendix~\ref{app:ablation} shows that reversing sentence polarity is
consistently more effective than only adding or removing negation cues.
Interestingly, 
pre-training on only one type of instances (add or remove negation) 
also statistically significantly outperforms the \hl{off-the-shelf} models for all models other than {BERT-large}.
We also observe that pre-training with the subset of the data that includes only adding negation
is more beneficial than pre-training with the subset of the data that includes only removing negation.

\subsection{Results on the Other Corpora}
\paragraph{NLI Corpora.}
Table~\ref{tab:nliresults} presents the results on RTE, SNLI, and MNLI.
Pre-training does not substantially change the accuracy on all the original instances. 
They are all within -1\% to +1.8\% of the \hl{off-the-shelf} models other than a 2.5\%-7.9\% improvement on RTE with {RoBERTa-base}.
This suggests that our models still perform as good as the \hl{off-the-shelf} models when negation is not present.
\hl{Interestingly, joint training on both tasks does not improve the performance over pre-training on NSP for any of the models other than RoBERTa-base.} \tbd{This sentence used to be in the last paragraph, now I moved it here.}

According to \citet{hossain-etal-2020-analysis}, 
the original \hl{development} sets contain too few negated instances to assess model performance on negation. 
Hence, we focus on their newly created negated pairs. 
See Appendix~\ref{app:nliresults} for detailed results on each pair type.

Further pre-training {BERT} models on any of the tasks consistently outperforms the \hl{off-the-shelf} models on the new negated pairs
(RTE: 5.1\%-25.5\%, SNLI: 1.9\%-6.1\%, MNLI: 0-1.6\%). 
Importantly, the models further pre-trained on NSP outperform the previous work of 
\citet{hosseini-etal-2021-understanding} and \citet{singh-etal-2023-nlms} 
by 0\%-4.2\% and 1\%-4.7\%, respectively.
{RoBERTa} shows limited improvements from further pre-training compared to {BERT} models. 
The base model benefits more from jointly pre-training, 
outperforming the \hl{off-the-shelf} model and the previous work. 
The large model, however, benefits more from pre-training on NSP but does not outperform the previous work on RTE.
Pre-training on NSPP consistently improves performance across all models, except for {RoBERTa-large}.
However, other than a 3.5\% improvement on SNLI with {BERT-base},
it is either less effective or within +1\% of the models pre-trained on NSP.

\paragraph{NLU Corpora.}

Table~\ref{tab:nluresults} presents the results on the validation sets of natural language understanding tasks. 
Following prior work, we report macro-averaged F1 scores on the validation sets as some test labels are not publicly available. 
The results demonstrate that further pre-training consistently improves performance on instances containing negation or, at worst, causes a negligible decline 
(a marginal 0.01\% decrease with {RoBERTa-base} on WSC.)
On average, pre-training yields a 3.11\% improvement across all tasks.
Notably, the most significant improvements are observed on WiC with base models
(achieving an average increase of 7.5\%) and on WSC with large models (performance improves by 6.0\% on average.) 
Importantly, all models pre-trained on NSP or NSPP outperform off-the-shelf versions on important instances in QNLI, 
with the only exceptions being {BERT-large} pre-trained on either NSPP or NSP.

\paragraph{LAMA and LAMA-Neg.}
Table~\ref{tab:lama-neg} presents the results on LAMA-Neg.
It is important to note that 
\hl{achieving 0\% error rate on}
the LAMA-Neg task only requires a model to behave randomly in the presence of negation. 
Therefore, while low top-1 mean error rates can serve as a sanity check, 
they do not reliably indicate the ability of the models to reason over negation.
Nevertheless, 
our models outperform the \hl{off-the-shelf} models on LAMA-Neg, 
with error rates reduced by 0.09 to 19.96 points (6.19 on average). 
Table~\ref{tab:ori-lama-results} presents the mean precision at $k = 1$ on the original LAMA dataset. 
Crucially, except for {RoBERTa} models jointly pre-trained on both NSP and NSPP, 
all other models remain within $\pm$1.65\% of the vanilla models. 
Most notably, models pre-trained on NSP and NSPP consistently outperform the vanilla models on {SQuAD} by 0.33\%-1.32\%,
with the exception of {BERT-base} pre-trained on NSPP, which performs 0.32\% worse.
It is important to note that LAMA does not contain negated instances.
Therefore, improvements are not necessarily expected. 
However, the fact that the models remain within $\pm$1.65\% of the vanilla models, 
coupled with the substantial improvements on LAMA-neg (Table~\ref{tab:lama-neg}) 
and other corpora, 
demonstrates that the models achieve more robustness to negation while maintaining competitive performance on inputs without negation.
\begin{table}[t]
\centering
\input{tables/lama-neg.tex}
\caption{
    We report the mean top 1 error rate for negated LAMA queries. 
    The lower the error rate, the better the model.
    All our models outperform the \hl{off-the-shelf} models.
    \label{tab:lama-neg}
}
\end{table}

%% file: tables/nliresults.tex
\newcommand{\sig}{$^{\ast}$}
\setlength{\tabcolsep}{0.06in}
\footnotesize
\begin{tabular}{l r@{\hspace{0.15in}} rrr r@{\hspace{0.10in}} rrr r@{\hspace{0.10in}} rrr}
\toprule 
&& \multicolumn{3}{c}{\textbf{RTE}} && \multicolumn{3}{c}{\textbf{SNLI}} && \multicolumn{3}{c}{\textbf{MNLI}} \\ \cmidrule{3-5} \cmidrule{7-9} \cmidrule{11-13}
&& \multicolumn{2}{c}{Original} & \multirow[c]{2}{*}{Negated} && \multicolumn{2}{c}{Original} & \multirow[c]{2}{*}{Negated} && \multicolumn{2}{c}{Original} & \multirow[c]{2}{*}{Negated} \\ \cmidrule{3-4} \cmidrule{7-8} \cmidrule{11-12}
&& dev & $\text{dev}_{\text{neg}}$ && & dev & $\text{dev}_{\text{neg}}$ && & dev & $\text{dev}_{\text{neg}}$ \\
\midrule
{BERT-base} && 64.6 & 53.7 & 52.7 && 89.9 & 89.4 & 44.8 && 84.5 & 84.3 & 63.5\\ 
~~~~+ NSPP && 66.4 & 61.0 & 60.7 && 90.9 & 91.7 & \textbf{50.9} && 84.6 & 84.7 & 63.8\\ 
~~~~+ NSP && 65.0 & \textbf{61.0} & \textbf{74.5} && 90.8 & 90.3 & 47.4 && 84.5 & 84.5 & 65.0\\ 
~~~~+ NSPP + NSP && 65.0 & 58.5 & 57.8 && \textbf{90.9} & \textbf{91.7} & 47.4 && \textbf{84.6} & \textbf{84.7} & \textbf{65.1}\\ 
{NBERT-base} \cite{singh-etal-2023-nlms} && \textbf{71.1} & 56.1 & 71.1 && 90.6 & 90.7 & 44.8 && 83.4 & 82.8 & 63.8 \\
{BERTNOT} \cite{hosseini-etal-2021-understanding} && 70.0 & n/a & 74.5 && 89.0 & n/a & 46.0 && 84.3 & n/a & 60.9 \\
\midrule
{BERT-large} && 67.5 & 56.1 & 53.4 && 91.6 & \textbf{92.6} & 50.0 && 86.2 & 86.1 & 67.5\\ 
~~~~+ NSPP && 67.1 & 56.1 & 69.1 && 91.5 & 91.2 & 51.9 && 86.3 & 86.3 & 67.9\\ 
~~~~+ NSP && 66.5 & \textbf{70.7} & \textbf{78.9} && 91.4 & 92.2 & \textbf{52.7} && 86.1 & 86.3 & 67.7\\ 
~~~~+ NSPP + NSP && 67.5 & 61.0 & 59.4 && 91.4 & 91.7 & 52.7 && \textbf{86.5} & \textbf{86.5} & \textbf{67.9}\\ 
{NBERT-large} \cite{singh-etal-2023-nlms} && \textbf{72.2} & 51.2 & 74.2 && \textbf{91.6} & 91.2 & 50.3 && 65.4 & 70.0 & 66.7 \\

\midrule
\midrule
{RoBERTa-base}   &&  70.8  &  61.0  &  58.7  &&  \textbf{92.1}  &  \textbf{91.7}  &  55.2  &&  87.8  &  88.0  &  67.1  \\ 
~~~~+ NSPP &&  73.3 & 73.2 & 67.7  &&  91.9  &  90.3  &  \textbf{55.5}  &&  87.4  &  87.6  &  67.5  \\ 
~~~~+ NSP &&  78.0 & 73.2 & 78.7  &&  91.7  &  90.8  &  54.6  &&  87.9  &  \textbf{88.4}  &  66.5  \\ 
~~~~+ NSPP + NSP &&  \textbf{78.7} & 75.6 & \textbf{81.0}  &&  91.9  &  90.8  &  55.4  &&  87.8  &  88.1  &  \textbf{68.1}  \\ 
{NRoBERTa-base} \cite{singh-etal-2023-nlms} && 73.5 & \textbf{79.1} & 79.0 && 91.0 & 91.3 & 50.8 && \textbf{88.0} & 87.2 & 66.2\\
\midrule
{RoBERTa-large}  &&  83.4  &  \textbf{92.7}  &  84.7  &&  \textbf{93.1}  &  93.1  &  56.0  &&  \textbf{90.6}  &  \textbf{90.7} &  69.9  \\ 
~~~~+ NSPP &&  83.8  &  87.8  &  81.1  &&  92.8  &  93.1  &  53.6  &&  89.8  &  90.0  &  69.7  \\ 
~~~~+ NSP &&  84.5  &  85.4  &  87.2  &&  93.0  &  92.6  &  \textbf{56.5}  &&  90.2  &  90.2  &  \textbf{69.9}  \\ 
~~~~+ NSP + NSPP &&  83.4  &  87.8  &  75.4  &&  93.0  &  \textbf{93.5}  &  56.1  &&  90.3  &  90.2  &  69.7  \\ 
{NRoBERTa-large}  \cite{singh-etal-2023-nlms} && \textbf{88.4} & 80.4 & \textbf{88.4} && 91.5 & 92.3 & 56.5 && 89.0 & 89.2 & 69.6 \\
\bottomrule
\end{tabular}

%% file: tables/nluresults.tex
\newcommand{\sig}{$^{\ast}$}
\setlength{\tabcolsep}{0.04in}
\small
\begin{tabular}{l r@{\hspace{0.15in}} rrrrr r@{\hspace{0.15in}} rrr r@{\hspace{0.15in}} rrr}
\toprule
&& \multicolumn{5}{c}{\textbf{QNLI}} && \multicolumn{3}{c}{\textbf{WiC}} && \multicolumn{3}{c}{\textbf{WSC}} \\ \cmidrule{3-7} \cmidrule{9-11} \cmidrule{13-15}
&& \multirow[c]{2}{*}{All} & \multirow[c]{2}{*}{w/o neg} & \multicolumn{3}{c}{w/ neg} && \multirow[c]{2}{*}{All} & \multirow[c]{2}{*}{w/o neg} & \multirow[c]{2}{*}{\ w/ neg} && \multirow[c]{2}{*}{All} & \multirow[c]{2}{*}{w/o neg} & \multirow[c]{2}{*}{\ w/ neg} \\ 
\cmidrule{5-7}
&& & & \ \ \ All & \ \ imp. & unimp. & & & & & & & \\  \midrule
{BERT-base} && 0.88 & 0.88 & 0.88 & 0.65 & 0.89 && 0.69 & 0.71 & 0.59 && 0.52 & 0.44 & 0.60 \\
~~~~+ NSPP && 0.87 & 0.87 & 0.88 & \textbf{0.90} & 0.84 && \textbf{0.71} & \textbf{0.71} & 0.68 && 0.53 & 0.42 & 0.63 \\
~~~~+ NSP  && \textbf{0.91} & \textbf{0.91} & 0.89 & 0.85 & 0.89 && 0.70 & 0.71 & 0.67 && \textbf{0.56} & \textbf{0.48} & \textbf{0.63} \\
~~~~+ NSPP + NSP && 0.91 & 0.91 & \textbf{0.90} & 0.85 & \textbf{0.90} && 0.70 & 0.71 & \textbf{0.68} && 0.52 & 0.42 & 0.62 \\
\midrule
{BERT-large} && 0.89 & 0.90 & 0.87 & \textbf{0.80} & 0.87 && 0.69 & 0.70 & 0.62 && 0.51 & 0.44 & 0.58 \\
~~~~+ NSPP && 0.92 & \textbf{0.93} & 0.90 & 0.70 & 0.90 && 0.69 & \textbf{0.70} & 0.63 && 0.56 & 0.48 & 0.63 \\
~~~~+ NSP  && \textbf{0.92} & 0.93 & \textbf{0.90} & 0.75 & \textbf{0.90} && \textbf{0.69} & 0.69 & \textbf{0.68} && 0.55 & 0.46 & \textbf{0.63} \\
~~~~+ NSPP + NSP && 0.90 & 0.90 & 0.88 & 0.80 & 0.88 && 0.68 & 0.69 & 0.65 && \textbf{0.60} & \textbf{0.54} & 0.65 \\
\midrule
\midrule
{RoBERTa-base} && 0.93 & 0.93 & 0.91 & 0.70 & 0.91 && 0.69 & \textbf{0.70} & 0.62 && 0.61 & 0.58 & 0.63 \\
~~~~+ NSPP && 0.93 & 0.93 & \textbf{0.92} & 0.75 & \textbf{0.92} && 0.68 & 0.68 & 0.67 && 0.63 & 0.65 & 0.62 \\
~~~~+ NSP  && \textbf{0.93} & \textbf{0.93} & 0.92 & \textbf{0.80} & 0.92 && \textbf{0.68} & 0.68 & 0.67 && 0.61 & 0.60 & \textbf{0.63} \\
~~~~+ NSPP + NSP && 0.92 & 0.93 & 0.91 & 0.75 & 0.92 && 0.68 & 0.68 & \textbf{0.71} && \textbf{0.63} & \textbf{0.66} & 0.62 \\
\midrule
{RoBERTa-large} && 0.93 & 0.93 & 0.92 & 0.78 & 0.92 && 0.71 & 0.71 & 0.66 && 0.69 & 0.67 & 0.71 \\
~~~~+ NSPP && 0.94 & 0.94 & 0.93 & 0.95 & 0.93 && 0.71 & 0.71 & 0.65 && 0.66 & 0.58 & 0.75 \\
~~~~+ NSP  && \textbf{0.94} & 0.94 & 0.93 & \textbf{0.95} & 0.93 && 0.71 & 0.72 & 0.67 && \textbf{0.77} & \textbf{0.79} & 0.75 \\
~~~~+ NSPP + NSP && 0.94 & \textbf{0.95} & \textbf{0.93} & 0.95 & \textbf{0.93} && 0.71 & 0.72 & 0.68 && 0.76 & 0.73 & \textbf{0.79} \\
~~~~w/ Affir. Interpret. && 0.94 & 0.94 & 0.92 & 0.89 & 0.92 && \textbf{0.73} & \textbf{0.73} & \textbf{0.70} && 0.71 & 0.68 & 0.75 \\

\bottomrule
\end{tabular}

%% file: tables/lama-neg.tex
\newcommand{\sig}{$^{\ast}$}
\setlength{\tabcolsep}{0.03in}
\small
\begin{tabular}{l cccc}
\toprule
LAMA-Neg& SQuAD & ConceptNet & TREx & GoogleRE \\
\midrule
{BERT-base} & 13.16 & 3.16 & 22.31 & 3.49 \\
~~~~ + NSPP & 7.89 & 2.34 & 6.93 & \textbf{0.37} \\
~~~~ + NSP & \textbf{5.26} & \textbf{1.49} &  \textbf{2.35} & 0.63 \\
~~~~ + NSPP + NSP & 9.87 & 1.81 & 3.08 & 0.65 \\
\midrule
{BERT-large} &  15.79 & 4.05 & 24.88 & 2.04 \\
~~~~ + NSPP & 10.86 & 2.43 & 9.23 & 1.95 \\
~~~~ + NSP & \textbf{7.57} & \textbf{2.27} & \textbf{5.71} & \textbf{0.03} \\
~~~~ + NSPP + NSP & 8.22 & 2.58 & 7.59 & 0.60 \\

\midrule
\midrule

{RoBERTa-base} & 9.21 & 2.87 & 11.36 & 0.65 \\
~~~~ + NSPP & 4.28 & \textbf{1.04} & 1.81 & 0.18 \\
~~~~ + NSP & 4.61 & 1.13 & 0.31 & 0.00 \\
~~~~ + NSPP + NSP & \textbf{3.29} & 1.33 & \textbf{0.21} & \textbf{0.00} \\
\midrule
{RoBERTa-large} & 13.49 & 4.63 & 14.32 & 1.03 \\
~~~~ + NSPP & \textbf{3.29} & \textbf{1.11} & 3.08 & 0.26 \\
~~~~ + NSP & 5.92 & 1.88 & \textbf{0.38} & \textbf{0.00} \\
~~~~ + NSPP + NSP & 4.28 & 1.11 & 0.78 & 0.00 \\
\bottomrule
\end{tabular}

%% file: tex/conclusion.tex
\section{Conclusion}
\begin{table}
    \centering
    \input{tables/lama.tex}
    \caption{
      We report the mean precision at $k = 1$ on the original LAMA dataset.
      The higher the precision, the better the model.
      Other than {RoBERTa} models jointly pre-trained on both tasks, 
      all our models are within $\pm$1.65\% of the vanilla models. 
      \label{tab:ori-lama-results}
    }
  \end{table}
\hl{In this work,
we proposed a self-supervised method to make language models more robust against negation.
We introduced two tasks to further pre-train LMs:
(a) the novel task of Next Sentence Polarity Prediction (NSPP)
and (b) a variation of the Next Sentence Prediction (NSP) task that involves reversing the polarity of the second sentence instead of selecting a random sentence.
Pre-training data for these tasks can be generated from any text corpus,
and the tasks do not require any human annotations.
Our experiments showed that further pre-training BERT and RoBERTa models
on these tasks consistently improves their performance on a range of negation-related benchmarks,
particularly on the challenging CondaQA corpus.
Notably, while both NSPP and NSP tasks are beneficial,
the NSP task yields consistently stronger improvements than the challenging NSPP task.
Perhaps surprisingly, training on both tasks jointly does not provide further benefits.
Future research could explore additional model architectures and pre-training corpora beyond Wikipedia
or include more contextual information (more than two sentences) in the pre-training tasks.}

%% file: tables/lama.tex
\newcommand{\sig}{$^{\ast}$}
\setlength{\tabcolsep}{0.03in}
\small
\begin{tabular}{l cccc}
\toprule
LAMA& SQuAD & ConceptNet & TREx & GoogleRE \\
\midrule

{BERT-base} & 13.11 & 12.71 & \textbf{29.48} & 9.25 \\
~~~~ + NSPP & 12.79 & \textbf{12.72} & 29.01 & \textbf{9.52} \\
~~~~ + NSP & \textbf{14.43} & 12.02 & 28.78 & 9.90 \\
~~~~ + NSPP + NSP & 14.10 & 12.53 & 29.32 & 8.92 \\
\midrule
{BERT-large} &  15.74 & 15.17 & \textbf{30.02} & 9.78 \\
~~~~ + NSPP & 16.72 & \textbf{15.38} & 29.75 & 9.85 \\
~~~~ + NSP & 17.05 & 14.40 & 29.00 & \textbf{10.03} \\
~~~~ + NSPP + NSP & \textbf{17.38} & 14.07 & 28.93 & 9.98 \\

\midrule
\midrule
{RoBERTa-base} & 9.18 & \textbf{14.77} & \textbf{11.93} & 2.77 \\
~~~~ + NSPP & 9.84 & 14.73 & 11.80 & \textbf{2.78} \\
~~~~ + NSP & \textbf{10.16} & 14.42 & 11.28 & 2.44 \\
~~~~ + NSPP + NSP & 8.20 & 12.06 & 6.76 & 2.36 \\
\midrule
{RoBERTa-large} & 13.44 & 18.28 & \textbf{15.48} & 2.24 \\
~~~~ + NSPP & 13.77 & 17.59 & 13.80 & \textbf{2.78} \\
~~~~ + NSP & \textbf{14.10} & \textbf{18.32} & 15.46 & 2.28 \\
~~~~ + NSPP + NSP & 7.54 & 17.34 & 3.68 & 0.64 \\
\bottomrule
\end{tabular}

%% file: apptex/rules.tex
\section{Details of Reversing Polarity}
\label{app:reversing}

\begin{table*}
    \centering
    \input{apptex/tables/aux.tex}
    \caption{
        Auxiliary verbs that we work with and how each negation cue is added to them.
        \label{fig:aux}
        }
\end{table*}

Table~\ref{fig:aux} shows the auxiliary verbs that we work with 
and how we add each of the three negation cues 
(not, n't, never) to them.
Some auxiliary verbs do not have a corresponding form for adding \hl{``n't''}
such as \emph{having}, \emph{may} and \emph{might}.
If there are multiple auxiliary verbs for a single verb,
we only negate the one that is most commonly used.
For example, in the sentence \emph{She might have been sleeping when you called.}
we only negate \emph{have} and not \emph{might}: \emph{She might have not been sleeping when you called.}
The result is grammatically correct, but it may not be the most common form in English.

%% file: apptex/tables/aux.tex
\begin{tabular}{l@{\hspace{0.5in}} lll}
\toprule
\textbf{Aux.} & \textbf{not} & \textbf{n't} & \textbf{never} \\
\midrule
be & not be & - & never be \\
being & not being & - & never being \\
was & was not & wasn't & was never \\
is & is not & isn't & is never \\
were & were not & weren't & were never \\
have & have not & haven't & have never \\
having & not having & - & never having \\
had & had not & hadn't & had never \\
've & 've not & - & 've never \\
do & do not & don't & do never \\
does & does not & doesn't & does never \\
did & did not & didn't & did never \\
can & can not & can't & can never \\
could & could not & couldn't & could never \\
will & will not & won't & will never \\
'll & 'll not & - & 'll never \\
would & would not & wouldn't & would never \\
shall & shall not & shan't & shall never \\
should & should not & shouldn't & should never \\
must & must not & - & must never \\
may & may not & - & may never \\
might & might not & - & might never \\
\bottomrule    
\end{tabular}

%% file: apptex/llm.tex
\section{LLMs for Reversing Polarity}
\label{app:llm-reversing}

\begin{figure*}
\centering
\begin{tabularx}{\textwidth}{l X}
\toprule
Prompt: & Remove ``not'' from the following sentence: 
Large amounts of heat are wasted when the boiler is not insulated \\ 
Llama-2-7B: & I apologize, but I cannot provide an answer to that question as it does not make sense. Insulation is not a source of heat, and it is not possible for heat to be wasted through a lack of insulation. Additionally, the sentence is not coherent as boilers are not capable of producing heat. Could you please rephrase or provide more context for your question? I'll be happy to help as best I can. \\
\midrule
Prompt: & Remove any negation from the following sentence:
Large amounts of heat are wasted when the boiler is not insulated. \\
Llama-2-7B: & Thank you for reaching out! I'm here to help you in a safe and respectful manner. I understand that you want me to remove any negation from the sentence ``Large amounts of heat are wasted when the boiler is not insulated.'' However, I must point out that the sentence itself is factually incoherent as heat cannot be wasted, as it is a fundamental physical quantity that is always conserved.
Instead, I suggest rephrasing the sentence to something more accurate and coherent, such as: ``Proper insulation of the boiler can help reduce heat loss and save energy.'' This way, we can provide a helpful and informative answer without promoting false or misleading information. Is there anything else I can assist you with? \\
\midrule
Prompt: & Simply remove any negation cues from the following sentence and fix the grammar. Do not change anything else. \\ 
 & Large amounts of heat are wasted when the boiler is not insulated. \\
Llama-2-7B: & Sure, I'd be happy to help! Here's the revised sentence without any negation cues and with corrected grammar: \\
& Large amounts of heat are wasted when the boiler is not insulated. \\
\bottomrule
\end{tabularx}
\caption{
        Examples of Llama-2-7B failing to remove the negation cue from a sentence.
        The model resists removing the negation cue, 
        arguing that the sentence is factually incorrect or incoherent.
        In the last example, 
        the model returns the original sentence without any changes claiming that it has removed the negation cue and fixed the grammar.
        \label{tab:llama2-remove}
    }
\end{figure*}

\begin{figure*}
\centering
\begin{tabularx}{\textwidth}{l X}
\toprule
Prompt: & Please remove ``n't'' from the following sentence and fix the structure of the sentence. Do not try to fix the sentence in terms of factuality. According to Russel, the system can recognise 50 words and identifies the correct word 94.14\% of the time but also skips words that it can't identify 18\% of the time. \\
ChatGPT: & According to Russell, the system can recognize 50 words and identifies the correct word 94.14\% of the time, but also skips words that it cannot identify 18\% of the time. \\
\midrule
Prompt: & Please remove any negation cues from the following sentence and fix the structure of the sentence. Do not try to fix the sentence in terms of factuality. According to Russel, the system can recognise 50 words and identifies the correct word 94.14\% of the time but also skips words that it can't identify 18\% of the time. \\
ChatGPT: & According to Russel, the system recognizes 50 words, identifies the correct word 94.14\% of the time, and skips words it cannot identify 18\% of the time. \\
\bottomrule
\end{tabularx}
\caption{
        Examples of prompting ChatGPT to remove negation cues from a sentence.
        In the first example, the model replaces the negation cue ``can't'' with ``cannot''. 
        Updating the prompt and asking the model to remove any negation cues rather than specifically ``n't'' results in the same problem.
        \label{tab:chatgpt-remove}
    }
\end{figure*}

\begin{figure*}
\centering
\begin{tabularx}{\textwidth}{l X}
\toprule
Prompt: & Add the negation cue ``not'' to the following sentence:
Late at night, Tod sneaks over to visit Copper. \\
Llama-2-7B: & I'm glad you're here to ask me questions! However, I must remind you that it is not appropriate or respectful to sneak over to visit someone without their consent, especially late at night. It's important to prioritize people's privacy and personal boundaries. Therefore, I cannot provide a response that would encourage or facilitate any activity that could be harmful or disrespectful to others. Is there anything else I can help you with? \\
\midrule
Prompt: & Add the negation cue ``not'' to the main verb of the following sentence. You do not need to worry about the factuality or appropriateness. This is a grammar exercise.
Late at night, Tod sneaks over to visit Copper. \\ 
Llama-2-7B: &  Not late at night, Tod sneaks over to visit Copper. \\
\bottomrule
\end{tabularx}
\caption{
        An example of Llama-2-7B adding the negation cue ``not'' to a sentence.
        The model resists adding the negation cue, arguing that the sentence is inappropriate or disrespectful.
        When asked to add the negation cue without considering the appropriateness or factuality of the sentence and focusing on grammar, 
        the model adds the negation cue to the beginning of the sentence instead of the main verb.
        \label{tab:llama2-add}
    }
\end{figure*}

We attempted to use state-of-the-art LLMs to reverse the polarity of sentences.
Figure~\ref{tab:llama2-remove} shows examples of Llama-2-7B failing to remove the negation cue from a sentence.
The model points out that the sentence without the negation cue is factually incorrect or incoherent.
Asking the model to focus only on the grammar and remove the negation cue results in the model returning the original sentence without any changes.
We did not experiment with larger versions of Llama-2 such as 13B or 70B due to the high computational cost and time required to run the models 
given the large number of sentences we need to reverse the polarity of (We have $\approx$ 12.8M sentences in our dataset, 
of which we used 500K or 1M sentences for pre-training {RoBERTa}). 
Our observations suggest that ChatGPT is better at reversing the polarity of sentences,
however, other than being expensive, it fails in some cases as well (Figure~\ref{tab:chatgpt-remove}).
Adding negation is also challenging for Llama-2-7B (Figure~\ref{tab:llama2-add}).

%% file: apptex/condaqa_example.tex
\section{CondaQA Example}
\label{sec:condaqaexample}
\begin{figure}
  \small
  \begin{tabularx}{0.48 \textwidth}{lX}
      \toprule
      \textbf{Type} & \textbf{Example} \\
      \midrule
      Original & He didn't go to the store, but he went to the park. \\
      Paraphrase & He went to the park but not the store. \\
      Scope & He went to the store, but he didn't go to the park. \\
      Affirmation & He went to the store and the park. \\
      \bottomrule
  \end{tabularx}
  \caption{
      Three types of edits in CondaQA are applied to an example sentence.
      \label{fig:condaqaeditexample}
  }
\end{figure}

\begin{figure*}
    \small
      \begin{tabularx}{\textwidth}{p{.9in}X} 
        \toprule
        {Original Passage:} & 33\% of the faculty are members of the National Academy of Science or Engineering and/or fellows of the American Academy of Arts and Sciences. This is the highest percentage of any faculty in the country with the exception of the graduate institution Rockefeller University. \\ \addlinespace
        {Original Sentence (with Negation):} & This is the highest percentage of any faculty in the country \textbf{with the exception of} the graduate institution Rockefeller University. \\ \addlinespace
        {Negation Cue:} & with the exception of \\ \addlinespace
        {Question:} & Are the majority of faculty at any school other than Rockefeller University members of the National Academy of Science or Engineering? \\ \addlinespace
        \midrule
        Paraphrase Edit: &  33\% of the faculty are members of the National Academy of Science or Engineering and/or fellows of the American Academy of Arts and Sciences. This is the highest percentage of any faculty in the country \emph{other than} the graduate institution Rockefeller University. \\ \addlinespace
        \midrule
        Scope Edit: &  33\% of the faculty are \emph{not} members of the National Academy of Science or Engineering and/or fellows of the American Academy of Arts and Sciences. This is the highest percentage of any faculty in the country with the exception of the graduate institution Rockefeller University \\ \addlinespace
        \midrule
        Affirmation Edit: &  33\% of the faculty are members of the National Academy of Science or Engineering and/or fellows of the American Academy of Arts and Sciences. This is the highest percentage of any faculty in the country \emph{including} the graduate institution Rockefeller University. \\ \addlinespace
        \midrule
      \end{tabularx}
      \begin{tabularx}{\textwidth}{l@{\hspace{0.28 \textwidth}}l}
        \midrule
            Input & Answer \\
        \midrule
            Question + Original Passage & No \\
            Question + Paraphrase Edit & No \\
            Question + Scope Edit & Yes \\
            Question + Affirmation Edit & No \\
        \bottomrule
      \end{tabularx}
        \caption{
          \label{fig:condaqasample}
          An example from CondaQA.
          The original passage contains a sentence with negation.
          The crowdworker makes three edits to the passage (paraphrase, scope, and affirmation edits) to create the edited passage.
          The question (also written by the crowdworker) asks about the majority of faculty (more than 50\%) at any school other than Rockefeller University.
          Changing the scope of negation changes the answer to the question from \emph{No} to \emph{Yes}.
        }
\end{figure*}
Figure~\ref{fig:condaqaeditexample} shows an example sentence with the three types of edits.
We also provide an example from CondaQA in Figure~\ref{fig:condaqasample}.
The original passage has been selected from the English Wikipedia and contains a sentence with negation.
Three edits are made to the passage to create the edited passage: 
a paraphrase edit (i.e. \emph{rewriting the sentence}), a scope edit (i.e. \emph{changing the scope of negation}), and an affirmation edit (i.e. \emph{undoing negation}).
The question is answered based on the original and edited passages (a group). 
A model needs to answer all the questions in a group correctly to achieve group consistency. 

%% file: apptex/nlunli_example.tex
\section{NLU and NLI Corpora}
\label{sec:nlunlicorpora}
\begin{figure*}
\small
\centering
    \begin{tabularx}{\textwidth}{lXp{1.25in}} 
        \toprule
        & Input & Output \\
        \midrule
        Natural Language Inference \\
        ~~~~QNLI & When was the last time San Francisco hosted a Super Bowl? & \multirow{2}{1.25in}{Not Entailment (i.e., question is not answered)} \\
        & The South Florida/Miami area has previously hosted the event 10 times (tied for most with New Orleans), with the most recent one being Super Bowl XLIV in 2010. \\
        ~~~~MNLI & $\text{T}_\text{neg}$: His knees were not bent. & Contradiction \\
        & H: He bent his legs. & \\
        ~~~~RTE & $\text{T}_\text{neg}$: Green cards are not becoming more difficult to obtain. $\text{H}_\text{neg}$: Green card is not now difficult to receive. & Entailment \\
        ~~~~SNLI & T: A very thin, black dog running in a field.  & Entailment \\
        & $\text{H}_\text{neg}$: The dog is not in the house.& \\   
        \midrule
        Word Sense Disambiguation \\
        ~~~~WiC & A \emph{check} on its dependability under stress. & Not same meaning \\
        & He paid all his bills by \emph{check}. \\
        \midrule
        Coreference Resolution \\
        ~~~~WSC & Sid explained his theory to \emph{Mark} but \emph{he} couldn't convince him. & Not coreferent \\
        \bottomrule
        \end{tabularx}
        \caption{
            \label{tab:nlu/icorpora}
            Examples from the NLU and NLI corpora used in this work.
            The corpora include natural language inference (NLI), word sense disambiguation (WiC), and coreference resolution (WSC) tasks.
            There are four NLI datasets: QNLI, MNLI, RTE, and SNLI.
            Examples for MNLI, RTE, and SNLI are selected from the new instances with negation created by \citet{hossain-etal-2020-analysis}.
        }
\end{figure*}
QNLI~\cite{rajpurkar-etal-2016-squad} is a natural language inference dataset 
created from the Stanford Question Answering Dataset (SQuAD)~\cite{rajpurkar-etal-2016-squad}.
It contains questions and sentences that are answers to other questions from SQuAD.
The task is to determine whether the context sentence contains the answer to the question.
WiC~\cite{pilehvar-camacho-collados-2019-wic} is a word sense disambiguation dataset.
It contains sentence pairs where a word can have the same or different meanings in the two sentences.
The task is to determine whether the word has the same meaning in the two sentences.
WSC~\cite{levesque_winograd_2012} is a coreference resolution dataset.
It contains sentences where a pronoun can refer to different entities in the sentence.
The task is to determine whether a pronoun and a noun phrase are co-referential.

We present examples from the NLU and NLI corpora used in this work in Figure~\ref{tab:nlu/icorpora}. 
The examples are from the development sets of the corpora 
other than the examples from the new instances with negation created by \citet{hossain-etal-2020-analysis} for MNLI, RTE, and SNLI.

%% file: apptex/lama.tex
\section{LAMA and LAMA-Neg}
\label{app:lama}

The LAMA (Language Model Analysis) dataset, introduced by ~\citet{petroni-etal-2019-language},
serves as a probe for evaluating the factual and commonsense knowledge embedded within pre-trained language models. 
It consists of facts structured as subject-relation-object triples or question-answer pairs, 
which are transformed into cloze-style sentences to test whether language models can predict the correct masked token. 
LAMA encompasses various knowledge sources covering a range of factual and commonsense relations.

SQuAD~\cite{rajpurkar-etal-2016-squad} the ability of models to answer natural language questions by transforming them into cloze-style sentences.
Google-RE\footnote{\footnotesize {code.google.com/archive/p/relation-extraction-corpus/}} 
assesses the ability to retrieve facts about dates, places of birth, and places of death. 
T-REx~\cite{elsahar-etal-2018-rex}, 
which contains a larger set of relations derived from Wikidata, challenges the models on their ability to generalize across a broader range of facts, 
although some noise exists due to the automatic alignment to Wikipedia. 
ConceptNet~\cite{speer-havasi-2012-representing}, 
on the other hand, tests commonsense reasoning by providing facts about everyday concepts and relationships.

LAMA-Neg, introduced by \citet{kassner-schutze-2020-negated},
extends the LAMA dataset by incorporating negated instances to evaluate the ability of pre-trained language models to handle negation. 
The dataset transforms positive cloze statements from LAMA into their negated counterparts by inserting negation cues such as “not” 
(e.g., “Einstein was born in [MASK]” becomes “Einstein was not born in [MASK]”). 
This addition allows them to probe whether models can correctly differentiate between positive and negative factual assertions.
The model is expected to predict any token other than the original masked token in the negated variant.

%% file: apptex/hyper.tex
\section{Implementation Details}
\label{app:hyperparameters}

\begin{table}
\centering
\begin{tabular}{l r}
\toprule
Training Set Size & Batch Size \\
\midrule
1k & 32 \\
2k & 32 \\
5k & 32 \\
10k & 32 \\
20k & 32 \\
50k & 64 \\
100k & 128 \\
200k & 256 \\
500k & 512 \\
1M & 512 \\
\bottomrule
\end{tabular}
\caption{
    Training set sizes and batch sizes used for pre-training.
    The batch size only changes for the training set sizes, not the task we are pre-training on.
    The learning rate is set to $1e-6$ for all training set sizes.
    \label{tab:trainingsetsize}
}

\end{table}

\begin{table*}
\centering
\begin{tabular}{l rrrr}
\toprule
& \multicolumn{1}{c}{Vanilla} & \multicolumn{1}{c}{NSPP} & \multicolumn{1}{c}{NSP} & \multicolumn{1}{c}{NSPP+NSP} \\
\midrule
{RoBEBTa-base} & \\
~~~~CondaQA & $5e-6(16)$ & $1e-5(8)$ & $5e-6(8)$ & $1e-5(16)$ \\
~~~~QNLI & $1e-5(16)$ & $5e-6(16)$ & $1e-5(16)$ & $1e-5(16)$ \\
~~~~WiC & $5e-5(16)$ & $1e-5(16)$ & $1e-5(16)$ & $1e-5(16)$ \\
~~~~WSC & $1e-6(16)$ & $1e-4(16)$ & $1e-6(16)$ & $1e-4(16)$ \\
~~~~RTE & $2e-5(32)$ & $2e-5(32)$ & $2e-5(32)$ & $2e-5(32)$ \\
~~~~SNLI & $1e-5(32)$ & $1e-5(32)$ & $1e-5(32)$ & $1e-5(32)$ \\
~~~~MNLI & $2e-5(32)$ & $2e-5(32)$ & $2e-5(32)$ & $2e-5(32)$ \\
\midrule
{RoBERTa-large} & \\
~~~~CondaQA & $1e-5(16)$ & $1e-5(16)$ & $5e-6(8)$ & $5e-6(16)$ \\
~~~~QNLI & $1e-5(16)$ & $5e-6(16)$ & $1e-5(16)$ & $5e-6(16)$ \\
~~~~WiC & $5e-5(16)$ & $5e-6(16)$ & $1e-5(16)$ & $1e-5(16)$ \\
~~~~WSC & $1e-6(16)$ & $5e-5(16)$ & $1e-6(16)$ & $1e-5(16)$ \\
~~~~RTE & $2e-5(32)$ & $2e-5(32)$ & $2e-5(32)$ & $2e-5(32)$ \\
~~~~SNLI & $1e-5(32)$ & $1e-5(32)$ & $1e-5(32)$ & $1e-5(32)$ \\
~~~~MNLI & $2e-5(32)$ & $2e-5(32)$ & $2e-5(32)$ & $2e-5(32)$ \\
\midrule
\midrule
{BEBT-base} & \\
~~~~CondaQA & $1e-5(8)$ & $1e-5(8)$ & $1e-5(8)$ & $1e-5(8)$ \\
~~~~QNLI & $1e-4(16)$ & $1e-6(16)$ & $5e-5(16)$ & $1e-6(16)$ \\
~~~~WiC & $1e-5(16)$ & $5e-5(16)$ & $5e-5(16)$ & $1e-5(16)$ \\
~~~~WSC & $1e-4(16)$ & $5e-6(16)$ & $1e-5(16)$ & $5e-6(16)$ \\
~~~~RTE & $2e-5(8)$ & $2e-5(8)$ & $2e-5(8)$ & $2e-5(8)$ \\
~~~~SNLI & $1e-5(32)$ & $1e-5(32)$ & $1e-5(32)$ & $1e-5(32)$ \\
~~~~MNLI & $2e-5(32)$ & $2e-5(32)$ & $2e-5(32)$ & $2e-5(32)$ \\
\midrule
{BERT-large} & \\
~~~~CondaQA & $5e-6(8)$ & $1e-5(8)$ & $5e-6(8)$ & $1e-5(8)$ \\
~~~~QNLI & $1e-6(16)$ & $1e-5(16)$ & $1e-5(16)$ & $5e-6(16)$ \\
~~~~WiC & $1e-5(16)$ & $1e-5(16)$ & $5e-5(16)$ & $1e-5(16)$ \\
~~~~WSC & $1e-6(16)$ & $5e-6(16)$ & $5e-6(16)$ & $1e-5(16)$ \\
~~~~RTE & $2e-5(8)$ & $2e-5(8)$ & $2e-5(8)$ & $2e-5(8)$ \\
~~~~SNLI & $1e-5(32)$ & $1e-5(32)$ & $1e-5(32)$ & $1e-5(32)$ \\
~~~~MNLI & $2e-5(32)$ & $2e-5(32)$ & $2e-5(32)$ & $2e-5(32)$ \\
\bottomrule
\end{tabular}
\caption{
    The learning rates (and batch sizes) used for fine-tuning the pre-trained models on the training sets of the corpora we evaluate on.   
    \label{tab:finetuninghyperparameters} 
}
\end{table*}

Batch sizes used for pre-training the off-the-shelf {RoBERTa} and {BERT} models on different training set sizes 
are shown in Table~\ref{tab:trainingsetsize}.
We use a learning rate of $1e-6$ for all experiments.

To fine-tune the further pre-trained models on the CondaQA, NLI, and NLU tasks, we utilize the implementations 
provided by \citet{rezaei2024paraphrasing},
 \citet{hossain-etal-2020-analysis}, 
 and \citet{hossain-etal-2022-analysis}, respectively. 
For the NLU and NLI tasks, we use the formatted versions of the datasets available in the 
GLUE \cite{wang-etal-2018-glue} and SuperGLUE \cite{wang2020superglue} benchmarks.
Table~\ref{tab:finetuninghyperparameters} presents the hyperparameters used for fine-tuning the models on the downstream tasks.

For evaluation on LAMA and LAMA-Neg, we do not fine-tune the models, 
as these tasks lack dedicated training sets. 
Notably, since both LAMA and LAMA-Neg are masking tasks, 
it is crucial to prevent catastrophic forgetting during further pre-training. 
To address this, 
we apply Elastic Weight Consolidation (EWC) regularization \cite{ewcregularization} with a coefficient of $\lambda = 1e-3$. 

EWC helps retain the knowledge acquired from the original masked language modeling task 
by penalizing changes to weights deemed important for this task. 
This regularization ensures that while the model adapts to new tasks during pre-training, 
it preserves essential knowledge required for effective performance on the masking tasks. 
This strategy allows us to evaluate the models reliably on LAMA and LAMA-Neg without compromising their original capabilities.

%% file: apptex/ablation.tex
\section{Ablation Study on Reversing the Polarity}
\label{app:ablation}
\begin{table*}[h!]
    \centering
    \input{tables/condaqa-ablation.tex}
    \caption{
        Results on CondaQA test set for models pre-trained on NSP task with different subsets of the pre-training data.
        An asterisk ($^{\ast}$) indicates a statistically significant improvement (McNemar's test \cite{mcnemar1947note}, $p < 0.05$)
        over the vanilla transformer model.
        Pre-training on the original data consistently outperforms pre-training on subsets of the data.
        However, except for {BERT-large}, pre-training on only one type of instances (add or remove negation) still statistically significantly outperforms the vanilla models.
        \label{tab:condaqa-ablation}
    }
\end{table*}

Recall that our pre-training data includes reversing the polarity of sentences, 
i.e., half the time we add negation to the original data and half the time we remove negation from the original data.
In this section, we provide an ablation study to understand the effect of reversing the polarity of sentences on the performance of the models.
That is, 
we compare our models pre-trained on the original data (obtained by reversing the polarity of sentences) 
with models that have been pre-trained on subsets of the original data where we only add negation or only remove negation.

However,
we need to be careful about the tasks we are evaluating.
Recall that we have two tasks: Next Sentence Polarity Prediction (NSPP) and Next Sentence Prediction (NSP).
Note that we do not reverse the polarity of sentences for the NSPP task and only use original instances.
Additionally, when jointly pre-training on both tasks (NSPP + NSP), NSPP requires both types of instances to be present in the training data.
Therefore, the comparison is only meaningful for the NSP task. 

Table~\ref{tab:condaqa-ablation} presents the results of models pre-trained on different subsets of the data on the CondaQA test set.
Pre-training on the original data consistently outperforms pre-training on subsets of the data by 0.3\% to 7.8\% in terms of accuracy 
and 0.7\% to 12.9\% in terms of group consistency. 
Importantly, 
further pre-training even on a subset of the data still statistically significantly outperforms the vanilla models. 
The only exception is {BERT-large} where pre-training with removing negation leads to a marginally lower accuracy (0.6\%)
and pre-training with adding negation leads to a marginally higher accuracy (0.3\%) compared to the vanilla model.
This suggests that while reversing the polarity of sentences is the best strategy,
pre-training on only one type of instances (add or remove negation) still improves the performance of the models.

Moreover, pre-training with the subset of the data that includes only adding negation consistently outperforms pre-training with the subset of the data that includes only removing negation.
We suspect this is because adding negation is more likely to make the sentence an incoherent follow-up to the first sentence than removing negation is.
Therefore, the model learns more about negation when trained on instances with added negation than when trained on instances with removed negation.

%% file: tables/condaqa-ablation.tex
\newcommand{\sig}{$^{\ast}$}
\setlength{\tabcolsep}{0.07in}
\small
\begin{tabular}{l r c@{\hspace{0.10in}}  r@{}l r@{}l@{\hspace{0.07in}} r@{}l@{\hspace{0.07in}} r@{}l@{\hspace{0.07in}} r@{}l@{\hspace{0.07in}}   c@{\hspace{0.10in}}   r@{}l r@{}l r@{}l r@{}l}
\toprule
& & & \multicolumn{10}{c}{Accuracy} && \multicolumn{8}{c}{Group Consistency} \\
\cmidrule{4-12} \cmidrule{15-22}
& \# Params. && All && Ori. && Par. && Sco. && Aff. &&& All && Par. && Sco. && Aff. \\
\midrule
{BERT-base} & 110M && 45.4 && 44.8 && 44.7 && 45.0 && 47.3 &&& 3.4 && 40.5 && 15.2 && 14.1 \\
~~~~ + NSP (all data)&&& 47.8 &\sig& 48.8 &\sig& 49.4 &\sig& 45.9 && 46.9 &&& 3.6 && 44.2 && 16.8 && 13.7 \\
~~~~~~~~~~~~Only add negation &&& 47.5 &\sig& 50.2 &\sig& 50.7 &\sig& 47.3 && 41.5 &&& 1.3 && 48.7 && 16.0 && 9.3 \\
~~~~~~~~~~~~Only remove negation &&& 47.2 &\sig& 47.4 &\sig& 46.7 &\sig& 43.4 &\sig& 51.1 &&& 2.9 && 44.2 && 14.8 && 15.8 \\
\midrule
{BERT-large} & 336M && 48.5 && 49.4 && 49.6 && 47.0 && 47.8 &&& 3.6 && 45.7 && 16.5 && 13.6 \\
~~~~ + NSP (all data)&&&  51.9 &\sig& 53.3 &\sig& 51.0 && 50.5 &\sig& 52.7 &\sig&& 11.1 && 45.2 && 25.5 && 25.5 \\
~~~~~~~~~~~~Only add negation &&& 48.8 && 52.1 &\sig& 52.1 &\sig& 48.3 && 42.5 &\sig&& 3.1 && 49.2 && 18.5 && 12.6 \\
~~~~~~~~~~~~Only remove negation &&& 47.9 && 51.3 && 51.7 && 47.2 && 41.1 &\sig&& 3.7 && 47.9 && 18.1 && 12.8 \\
\midrule
\midrule
{RoBERTa-base} & 125M && 48.2 && 50.1 && 49.9 && 48.5 && 44.1 &&& 2.4 && 48.4 && 17.0 && 11.1 & \\
~~~~ + NSP (all data) &  &&  57.3 &\sig&  57.8 &\sig&  57.9 &\sig&  53.7 &\sig&  59.6 &\sig&&  17.6 &&  50.4 &&  30.8 &&  34.6 \\
~~~~~~~~~~~~Only add negation &&& 49.8 &\sig& 53.3 &\sig& 53.7 &\sig& 46.8 && 45.2 &&& 4.7 && 51.1 && 19.2 && 15.3 \\
~~~~~~~~~~~~Only remove negation &&& 49.5 &\sig& 52.0 && 52.5 &\sig& 48.3 && 44.9 &&& 4.8 && 50.1 && 18.6 && 13.9 \\
\midrule
{RoBERTa-large} & 355M && 64.9 && 66.6 && 66.3 && 60.7 && 65.6 &&& 29.6 && 61.9 && 41.4 && 45.8 \\
~~~~ + NSP (all data)&&&  68.0 &\sig&  71.1 &\sig&  70.2 &\sig&  63.3 &\sig&  67.1 &&&  34.0 &&  67.0 &&  46.6 &&  50.1 \\
~~~~~~~~~~~~Only add negation &&& 67.3 &\sig& 69.3 &\sig& 68.6 &\sig& 63.6 &\sig& 67.6 &&& 33.0 && 64.0 && 45.9 && 48.9 \\
~~~~~~~~~~~~Only remove negation &&& 67.0 &\sig& 68.9 &\sig& 69.0 &\sig& 62.9 && 66.7 &&& 32.5 && 64.7 && 45.1 && 47.1 \\
\bottomrule

\end{tabular}

%% file: apptex/nli-details.tex
\section{Detailed Results on NLI Tasks}
\label{app:nliresults}

\begin{table*}[h!]
    \centering
    \input{tables/nliresults-detailed.tex}
    \caption{
        Accuracies for each type of the new pairs containing negation for the natural language inference tasks from \citet{hossain-etal-2020-analysis}.
        T and H refer to the text and hypothesis sentences, respectively. We use $\text{T}_{\text{neg}}$ and $\text{H}_{\text{neg}}$ to denote the sentences with negation cues added to the main verb.
        \label{tab:nlidetailedresults}
    }
\end{table*}

\citet{hossain-etal-2020-analysis} created new pairs containing negation by adding negation to the premise or text of the original pairs in the validation sets of the natural language inference tasks.
Table~\ref{tab:nlidetailedresults} shows the accuracies of our models for each type of the new pairs containing negation.

%% file: tables/nliresults-detailed.tex
\newcommand{\sig}{$^{\ast}$}
\setlength{\tabcolsep}{0.03in}
\footnotesize
\begin{tabular}{l r@{\hspace{0.10in}} rrrr r@{\hspace{0.10in}} rrrr r@{\hspace{0.10in}} rrrr}
\toprule 
&& \multicolumn{4}{c}{\textbf{RTE}} && \multicolumn{4}{c}{\textbf{SNLI}} && \multicolumn{4}{c}{\textbf{MNLI}} \\ \cmidrule{3-6} \cmidrule{8-11} \cmidrule{13-16}
&& All & $\text{T}_{\text{neg}}$-H & T-$\text{H}_{\text{neg}}$ & $\text{T}_{\text{neg}}$-$\text{H}_{\text{neg}}$ && All & $\text{T}_{\text{neg}}$-H & T-$\text{H}_{\text{neg}}$ & $\text{T}_{\text{neg}}$-$\text{H}_{\text{neg}}$ && All & $\text{T}_{\text{neg}}$-H & T-$\text{H}_{\text{neg}}$ & $\text{T}_{\text{neg}}$-$\text{H}_{\text{neg}}$ \\
\midrule
{BERT-base} && 52.7 & 55.6 & 40.2 & 62.4 && 44.8 & 32.6 & 58.8 & 41.8 && 63.5 & 62.0 & 65.6 & 63.0\\ 
~~~~+ NSPP && 60.7 & 68.6 & 45.0 & 68.2 && 50.9 & 45.0 & 63.6 & 44.2 && 63.8 & 62.4 & 66.0 & 63.0\\ 
~~~~+ NSP && 74.5 & 77.2 & 78.4 & 67.8 && 47.4 & 43.2 & 61.0 & 38.0 && 65.0 & 63.6 & 67.0 & 64.4\\ 
~~~~+ NSPP + NSP && 57.8 & 65.4 & 40.4 & 67.6 && 47.4 & 41.4 & 62.0 & 38.8 && 65.1 & 63.8 & 66.8 & 64.6\\ 
{NBERT-bsae} && 71.1 & 72.8 & 84.6 & 56.0 && 44.8 & 38.0 & 60.6 & 35.8 && 63.8 & 64.2 & 65.5 & 61.4\\
\midrule
{BERT-large} && 53.4 & 57.0 & 40.8 & 62.4 && 50.0 & 41.2 & 62.2 & 46.6 && 67.5 & 65.2 & 71.6 & 65.8\\ 
~~~~+ NSPP && 69.1 & 77.8 & 63.0 & 66.6 && 51.9 & 43.4 & 64.6 & 47.8 && 67.9 & 65.6 & 73.2 & 64.8\\ 
~~~~+ NSP && 78.9 & 80.2 & 91.0 & 65.6 && 52.7 & 43.2 & 68.0 & 47.0 && 67.7 & 65.2 & 72.0 & 65.8\\ 
~~~~+ NSPP + NSP && 59.4 & 56.4 & 54.0 & 67.8 && 52.7 & 44.0 & 67.2 & 47.0 && 67.9 & 65.2 & 72.4 & 65.2\\ 
{NBERT-large} && 74.2 & 78.4 & 80.4 & 64.0 && 50.3 & 47.6 & 60.6 & 42.8 && 66.7 & 65.4 & 70.0 & 64.8\\
\midrule
\midrule
{RoBERTa-base} &&  58.7  &  56.6  &  53.6  &  66.0  &&  55.2  &  49.2  &  68.6  &  47.8  &&  67.1  &  65.4  &  71.0  &  65.0  \\ 
~~~~+ NSPP &&  67.7 & 79.2 & 56.0 & 67.8  &&  55.5  &  48.0  &  69.2  &  49.4  &&  67.5  &  65.0  &  71.0  &  66.6  \\ 
~~~~+ NSP && 78.7 & 86.6 & 78.2 & 71.4  &&  54.6  &  48.2  &  67.8  &  47.8  &&  66.5  &  64.6  &  70.2  &  64.6  \\ 
~~~~+ NSP + NSPP &&  81.0 & 88.0 & 83.6 & 71.4  &&  55.4  &  48.2  &  69.8  &  48.2  &&  68.1  &  66.0  &  72.4  &  66.0  \\
{NRoBERTa-base} && 79.0 & 80.0 & 91.0 & 66.2 && 50.8 & 42.8 & 65.4 & 44.2 && 66.2 & 65.4 & 69.2 & 64.2\\
\midrule
{RoBERTa-large}  &&  84.7  &  90.4  &  87.6  &  76.2  &&  56.0  &  51.4  &  69.4  &  47.2  &&  69.9  &  70.0  &  73.2  &  66.4  \\ 
~~~~+ NSPP &&  81.1  &  83.8  &  84.2  &  75.2  &&  53.6  &  48.2  &  64.8  &  47.8  &&  69.7  &  69.0  &  72.4  &  67.6  \\ 
~~~~+ NSP &&  87.2  &  91.0  &  90.8  &  79.8  &&  56.5  &  50.2  &  70.4  &  48.6  &&  69.9  &  68.6  &  74.2  &  66.8  \\ 
~~~~+ NSP + NSPP &&  75.4  &  90.0  &  70.0  &  67.8  &&  56.1  &  50.4  &  70.0  &  48.0  &&  69.7  &  69.6  &  73.0  &  66.4  \\
{NRoBERTa-large} && 88.4 & 93.4 & 90.6 & 81.2 && 56.5 & 50.4 & 70.4 & 48.8 && 69.6 & 67.4 & 74.2 & 67.2 \\

\bottomrule
\end{tabular}